\documentclass[acmsmall]{acmart}

\usepackage{bm}
\usepackage{multirow}

\newcounter{rown}
\newcommand{\rownum}{\stepcounter{rown}\arabic{rown}}
\AtBeginDocument{%
  }

\setcopyright{acmlicensed}
\copyrightyear{2024}
\acmYear{2024}
\acmDOI{XXXXXXX.XXXXXXX}

\acmJournal{CSUR}
\acmVolume{37}
\acmNumber{4}
\acmArticle{111}
\acmMonth{11}

\begin{document}

\title[WSSS with Image-level Labels: from Traditional Models to Foundation Models]{Weakly-Supervised Semantic Segmentation with Image-Level Labels: from Traditional Models to Foundation Models}

\author{Zhaozheng~Chen}
\affiliation{%
  \institution{Singapore Management University}
  \country{Singapore}}
\email{zzchen.2019@phdcs.smu.edu.sg}

\author{Qianru~Sun}
\affiliation{%
  \institution{Singapore Management University}
  \country{Singapore}}
\email{qianrusun@smu.edu.sg}

\renewcommand{\shortauthors}{Chen et al.}


\begin{abstract}
The rapid development of deep learning has driven significant progress in image semantic segmentation---a fundamental task in computer vision. Semantic segmentation algorithms often depend on the availability of pixel-level labels (i.e., masks of objects), which are expensive, time-consuming, and labor-intensive. %
Weakly-supervised semantic segmentation (WSSS) is an effective solution to avoid such labeling.
It utilizes only partial or incomplete annotations and provides a cost-effective alternative to fully-supervised semantic segmentation.
In this journal, our focus is on the WSSS with image-level labels, which is the most challenging form of WSSS.
Our work has two parts. First, we conduct a comprehensive survey on traditional methods, 
primarily focusing on those presented at premier research conferences. 
We categorize them into four groups based on where their methods operate: pixel-wise, image-wise, cross-image, and external data.
Second, we investigate the applicability of visual foundation models, such as the Segment Anything Model (SAM), in the context of WSSS.
We scrutinize SAM in two intriguing scenarios: text prompting and zero-shot learning. We provide insights into the potential and challenges of deploying visual foundational models for WSSS, facilitating future developments in this exciting research area.
Our code is provided at this \href{https://github.com/zhaozhengChen/SAM_WSSS}{link}.
\end{abstract}



\received{25 September 2023}
\received[revised]{1 Jul 2024}
\received[revised]{18 Sep 2024}
\received[accepted]{19 Nov 2024}

\maketitle

\section{Introduction}
\label{sec:intro}
Semantic segmentation serves as a fundamental task in the field of computer vision. This task involves labeling each pixel within an image with semantically meaningful labels corresponding to specific objects present. Semantic segmentation has broad applicability, with use cases ranging from autonomous driving and scene comprehension to medical image analysis. Enabling machines to extract rich semantic information from images allows them to understand the visual world like human perception. However, the high variability and complexity of real-world scenes~\cite{cityscape}, coupled with the requirement of extensive labeled data for training deep-learning models, make semantic segmentation a challenging task. To address these challenges, various approaches have been proposed, including fully convolutional networks~\cite{fcn}, encoder-decoder architectures~\cite{deeplabv2,deeplabv3+}, and attention mechanisms~\cite{se_ss}. These techniques have significantly advanced state-of-the-art semantic segmentation, making it a highly active and exciting research area.

Fully-supervised semantic segmentation requires a large number of labeled images for training. Unlike it, weakly-supervised semantic segmentation (WSSS) uses only partial or incomplete annotations to learn the segmentation task. This makes the weakly-supervised approach more feasible for real-world applications, where obtaining large amounts of fully labeled data can be prohibitively expensive or time-consuming. Weakly-supervised methods typically rely on various forms of supervision, such as image-level labels~\cite{sec,adv_erasing,advcam}, scribbles~\cite{scribble1,scribble2}, or bounding boxes~\cite{bbox1,bbox2,bbam}, to guide the segmentation process. Weakly-supervised techniques have shown remarkable progress in recent years, and they represent a promising direction for the future development of semantic segmentation algorithms. 

In this context, WSSS with image-level class labels is the most challenging and widely studied form of WSSS. In WSSS with image-level class labels, the only form of supervision provided is the class label for the entire image rather than for each individual pixel. The challenge is to use this limited information to learn the boundaries of objects and accurately segment them. To address this challenge, Class Activation Map (CAM)~\cite{cam} has emerged as a powerful technique in WSSS. CAM provides a way to visualize the areas of an image that are most relevant to a particular class without requiring pixel-level annotations. CAM is computed from a classification model by weighting the feature maps with the learned weights of the last fully connected layer, resulting in a heat map that highlights the most discriminative regions of an image. However, CAM often fails to capture the complete extent of an object, as it only highlights the most discriminative parts and leaves out other important regions. Most WSSS research focuses on generating more complete CAMs. We categorize these methods into four distinct groups, each defined by the level at which they operate:
\begin{itemize}
    \item Pixel-wise methods. These are methods that operate at the pixel level, employing strategies such as the usage of pixel-wise loss functions or the exploitation of pixel similarity and local patches to generate more accurate CAMs.
    \item Image-wise methods.  This category includes methods operating on a whole image level. Key methods encompass adversarial learning, context decoupling, consistency regularization, and the implementation of novel loss functions.
    \item Cross-image methods. Methods operate beyond a single image, extending their functions across pairs or groups of images. In some scenarios, these may cover the full extent of the dataset.
    \item Methods with external data. These are methods that utilize additional data sources beyond the training datasets, such as saliency maps and out-of-distribution data, to help the model better distinguish the co-occurring background cues.
\end{itemize}

In addition to these traditional methodologies in WSSS, our study also delves into the applicability and efficacy of recent foundation models. The foundation models, including GPT-3~\cite{gpt3}, CLIP~\cite{clip}, and SAM~\cite{sam}, have had a profound impact on both computer vision and natural language processing. This impact is largely attributed to their dependence on extensive data and the utilization of billions of model parameters. Among them, the Segment Anything Model (SAM)~\cite{sam} is specially crafted for the segmentation field. SAM introduces a new promptable segmentation task that supports various types of prompts, such as points, bounding boxes, and textual descriptions. It leverages a Transformer model~\cite{vit} trained on the extensive SA-1B dataset (comprising over 1 billion masks derived from 11 million images), which gives it the ability to handle a wide range of scenes and objects. SAM is remarkable for its capability to interpret diverse prompts and successively generate various object masks. 

In this survey, we assess the potential of SAM in WSSS by exploring two distinct settings: text input and zero-shot learning. In the text input setting, we first employ the Grounded DINO model~\cite{groundingdino} to generate bounding boxes of the target objects and then feed the bounding boxes and the image into SAM to yield the masks. In the zero-shot setting, where class labels are assumed to be absent (mirroring the validation process in WSSS), we first employ the Recognise Anything Model (RAM)~\cite{ram} to identify class labels. Subsequently, the same as the text input setting, the Grounded DINO is used to obtain the bounding boxes, and SAM is used to obtain masks.  A notable difference between the two settings lies in their subsequent steps. In the text input setting, it is necessary to train a segmentation model, such as DeepLabV2~\cite{deeplabv2}, to produce masks for the validation and test sets. In contrast, the zero-shot approach obviates the need for training an additional segmentation model.

\begin{figure}[ht]
\centering
\includegraphics[width=0.9\linewidth]{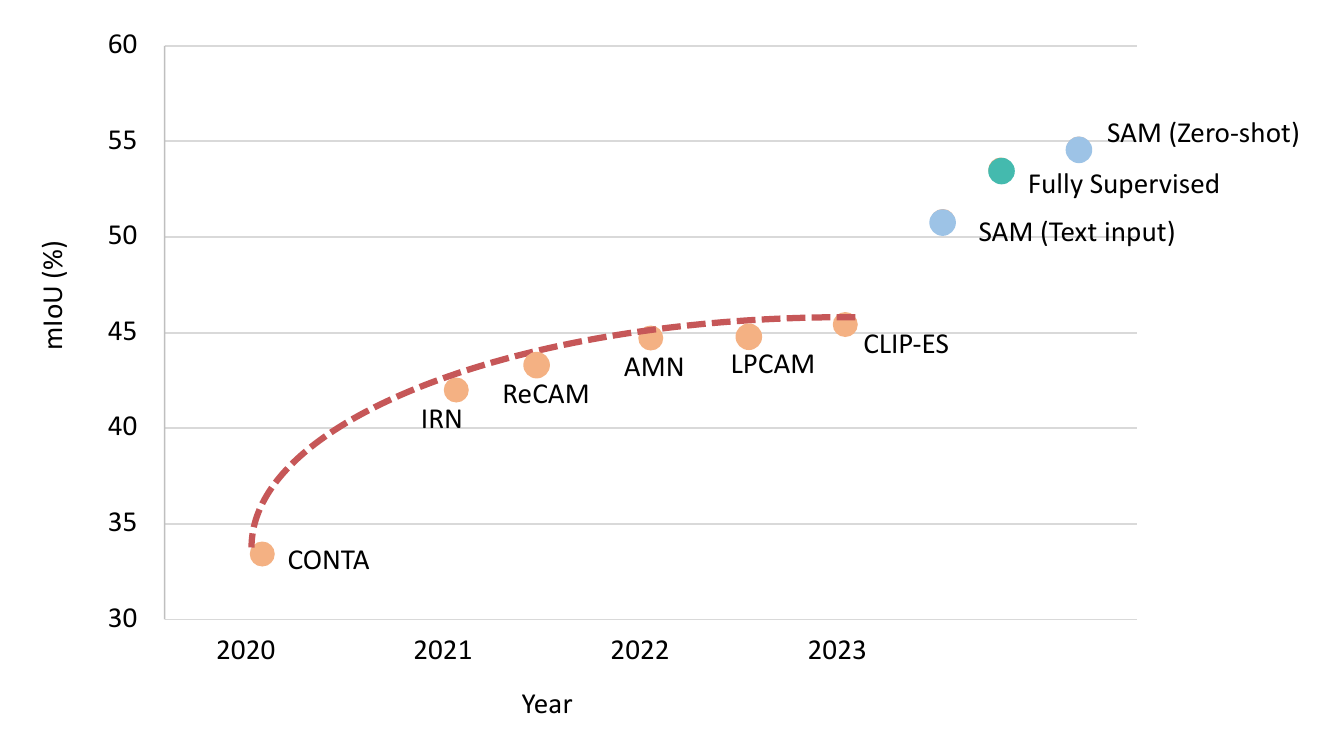}
\caption{The performance of recent WSSS works (CONTA~\cite{conta}, IRN~\cite{irn}, ReCAM~\cite{recam}, AMN~\cite{amn}, LPCAM~\cite{lpcam}, and CLIP-ES~\cite{clipes}) and an evaluation of foundation models on MS~COCO~\cite{mscoco} \texttt{val} set.}
\Description{
This figure compares the performance of recent weakly supervised semantic segmentation (WSSS) methods, including CONTA, IRN, ReCAM, AMN, LPCAM, and CLIP-ES, alongside foundation models such as SAM (evaluated in both zero-shot and text input settings) and a fully-supervised model on the MS COCO validation set. The x-axis denotes the publication year of the methods, spanning from 2020 to 2023, while the y-axis represents the mIoU performance as a percentage. The plot reveals a clear trend of increasing performance over time, with traditional methods reaching a performance plateau. In contrast, foundation models like SAM and the fully-supervised model demonstrate a substantial leap in mIoU scores, outperforming traditional WSSS approaches by a significant margin.
}
\label{fig:teaser}
\end{figure}
We compare the performance of the traditional methods and foundation models on the MS~COCO~\cite{mscoco} validation set. As shown in Fig.~\ref{fig:teaser},  the traditional methods reach a noticeable plateau in performance and the introduction of SAM has significantly enhanced the performance outcomes. In Section~\ref{sec:comparison}, we provide a comprehensive performance comparison of traditional methods and foundation models, offering insights into the potential and challenges of deploying foundational models in WSSS.

The paper is organized as follows: 
Section~\ref{sec:preliminary} introduces the preliminaries for the WSSS task.
Section~\ref{sec:tradition} introduces the traditional models in WSSS.
Section~\ref{sec:foundation} introduces the applicability of visual foundation models in WSSS.
Section~\ref{sec:comparison} provides a comprehensive comparison of the performance of the traditional models and the application of foundation models.
We conclude and discuss the future works in Section~\ref{sec:conclusion}.

\section{Preliminaries}
\label{sec:preliminary}

\subsection{Class Activation Map (CAM)}
\label{sec:cam}
Class Activation Map (CAM)~\cite{cam} is a simple yet effective technique employed to identify the regions within an image that a CNN leverages to identify a specific class shown in that image. It is calculated by multiplying the classifier weights with the image features. The specifics of how CAM is computed will be discussed in the subsequent discussion.

In the multi-label classification model, Global Average Pooling (GAP) is utilized, followed by a prediction layer. To compute the prediction loss on each training example, the Binary Cross-Entropy (BCE) function is employed, as detailed in the following formula:

{\small
\begin{equation}\label{eq:bce}
    \mathcal{L}_{bce}=-\frac{1}{K} \sum_{k=1}^{K} y \left[ k \right] \log \sigma\left(z[k]\right)+\left(1-y[k]\right) \log \left[1-\sigma\left(z[k]\right)\right],
\end{equation}}

\noindent
where \(z[k]\) denotes the prediction logit of the \(k\)-th class, \(\sigma(\cdot)\) is the sigmoid function, and \(K\) is the total number of foreground object classes (in the dataset). \(y[k]\in\{0,1\}\) is the image-level label for the \(k\)-th class, where 1 denotes the class is present in the image and 0 otherwise. Once the classification model converges, we feed the image $\bm{x}$ into it to extract the CAM of class $k$ appearing in $\bm{x}$:

\begin{equation} \label{equation:cam}
    \operatorname{CAM}_k(\bm{x})=\frac{\operatorname{ReLU}\left(\bm{A}_k\right)}{\max \left(\operatorname{ReLU}\left(\bm{A}_k\right)\right)}, \bm{A}_k=\mathbf{w}_{k}^{\top}f(\bm{x}),
\end{equation}
where \(\mathbf{w}_{k}\) denotes the classification weights (e.g., the FC layer of a ResNet) corresponding to the \(k\)-th class, and \(f(\bm{x})\) represents the feature maps of \(\bm{x}\) before the GAP.

As we mentioned in Section~\ref{sec:intro}, CAM often struggles to capture the complete object, instead focusing on the most distinctive parts. As such, a significant portion of the work in WSSS aims to tackle this issue, endeavoring to produce more complete CAMs.

\begin{figure}[ht]
\centering
\includegraphics[width=0.99\linewidth]{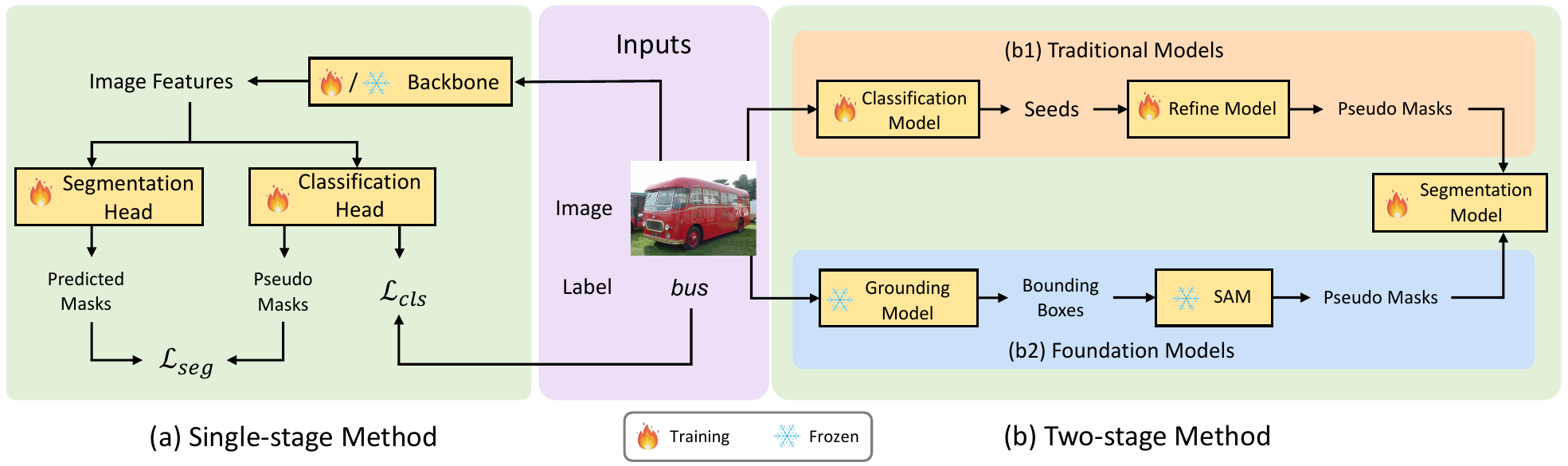}
\caption{The pipeline of existing methods in WSSS. Based on the number of stages, they can be divided into (a) single-stage and (b) two-stage methods. Based on the models used, they can be divided into traditional methods and foundation models.}
\Description{
This figure presents the pipeline of existing methods in weakly supervised semantic segmentation (WSSS). Panel (a) illustrates the single-stage method, where image features are extracted using a backbone model, which can be either trainable or frozen. These features are processed by a segmentation head to predict masks and a classification head to generate pseudo masks, with both outputs optimized using segmentation loss and classification loss, respectively. Panel (b) describes the two-stage method, divided into two categories: (b1) traditional models and (b2) foundation models. In (b1), a classification model generates seeds that are refined into pseudo masks by a refine model before being fed into a segmentation model. In (b2), a grounding model identifies bounding boxes, which are processed by SAM to produce pseudo masks.
}
\label{fig:pipeline}
\end{figure}
\subsection{Problem Statement}
Based on the number of stages, the WSSS works can be divided into single-stage (Fig.~\ref{fig:pipeline}~(a)) and two-stage methods (Fig.~\ref{fig:pipeline}~(b)). Single-stage methods~\cite{1stage,rrm,afa,tscd,toco,weclip} use a shared backbone to enable concurrent training for both the classification and segmentation tasks. The classification branch produces pseudo masks that are used to supervise the segmentation branch. This approach is jointly optimized end-to-end via a standard back-propagation algorithm during training. During inference, the classification head is discarded, and the segmentation branch is used to produce the mask. The two-stage method splits the process into two distinct phases. In the first stage, a classification model is trained to generate the seed areas of target classes from image-level labels. These initial seeds act as preliminary segmentation cues and are refined into pseudo masks to supervise the training of the segmentation model in the second stage. In the second stage, the final segmentation model (e.g., DeepLabV2~\cite{deeplabv2}) is then trained using these refined masks as pseudo ground truth. The two-stage methods can produce higher quality masks as they use a more focused refinement process and advanced segmentation model.

Based on the models used, they can be divided into traditional models as well as foundation models. The single-stage methods (Fig.~\ref{fig:pipeline}~(a)) can use traditional models (e.g. VGG~\cite{vgg}, ResNet~\cite{resnet}) or foundation models (e.g. CLIP~\cite{clip}) as the shared backbone. The two-stage methods can use traditional models (Fig.~\ref{fig:pipeline}~(b1)) or foundation models (Fig.~\ref{fig:pipeline}~(b2)) to produce pseudo masks to supervise the training of the segmentation model.
Fig.~\ref{fig:pipeline} (b1) illustrates the general pipeline for WSSS using traditional models. It begins with training a multi-label classification model using the images annotated with image-level class labels. Following this, they infer the class-specific seed areas for each image by applying the Class Activation Map (CAM)~\cite{cam} to the classification model. This results in a set of preliminary masks that undergo further refinement to produce the final pseudo masks. These pseudo masks then serve as the pseudo ground truth, enabling the training of a conventional fully supervised semantic segmentation model (e.g., DeepLabV2~\cite{deeplabv2}).
Fig.~\ref{fig:pipeline} (b2) shows the pipeline of applying foundation models in WSSS.
Given the images and their class labels, we first leverage a grounding model (e.g. Grounding DINO~\cite{groundingdino}) to generate bounding boxes using text prompts. Then, we feed these bounding boxes into SAM to produce corresponding segmentation masks. Similar to the traditional methods, we can train a fully supervised semantic segmentation model using the produced masks.

In this survey, we discuss those works from the perspective of traditional models and foundation models. We introduce the traditional model-based works and propose a new taxonomy to categorize them in Section~\ref{sec:tradition}. In Section~\ref{sec:foundation}, we explore the applicability of visual foundation models.

\subsection{Related Works}
To the best of our knowledge, there are only a few related survey papers~\cite{analysis_wsss,survey_label_efficient} on WSSS.
Chan~et~al.~\cite{analysis_wsss} focused on evaluating state-of-the-art weakly-supervised semantic segmentation methods (up until 2019) across diverse datasets, including those for natural scenes, histopathology, and satellite images.
Shen~et~al.~\cite{survey_label_efficient} provided a comprehensive review of label-efficient segmentation methods. They established a taxonomy classifying these methods based on the level of supervision, varying from no supervision to coarse, incomplete, and noisy supervision. Furthermore, they considered different types of segmentation problems like semantic segmentation, instance segmentation, and panoptic segmentation. In their work, WSSS methods (up until 2021) that utilize image-level class labels fall under the coarse supervision category, further subdivided into six distinct parts (i.e., seed area refinement by cross-label constraint, seed area refinement by cross-pixel similarity, seed area refinement by cross-view consistency, seed area refinement by cross-image relation, pseudo mask generation by cross-pixel similarity, and pseudo mask generation by cross-image relation).

In comparison, our paper offers a novel perspective by proposing a new taxonomy to categorize traditional WSSS methods with image-level class labels. We also take a step further by exploring the applicability and effectiveness of recent foundation models in the WSSS context, providing up-to-date insights into this rapidly developing field.
\section{Traditional Models}
\label{sec:tradition}
 
All the WSSS methods aim to produce high-quality pseudo masks from image-level labels. The non-CAM methods try to produce masks by constrained optimization~\cite{image_2_pixel,wsss_em,constrained}.
Pinheiro et~al.~\cite{image_2_pixel} first proposed using a CNN to learn features instead of designing task-specific features in the WSSS problem. The proposed method is based on an image classification model. It is trained by replacing the global max pooling layer with the Log-Sum-Exp~\cite{convex_optimization} layer, which is a smooth version and convex approximation of the max function. In this way, it casts the problem of segmentation into the problem of finding pixel-level labels from image-level labels.
Papandreou et~al.~\cite{wsss_em} developed an online Expectation-Maximization (EM) method for training the semantic segmentation model from image-level labels. The proposed algorithm alternates between estimating the latent pixel labels and optimizing the segmentation model parameters. It estimates the pixel labels by introducing an adaptive bias that boosts ground truth classes to be present and suppresses all others on the model's prediction.
CCNN~\cite{constrained} explores constrained optimization for WSSS. It models a distribution over latent ``ground truth'' labels while optimizing the CNN to follow this latent distribution as closely as possible. This allows them to enforce the constraints on the latent distribution instead of the model output. They applied four constraints (i.e., suppression, foreground, background, and size) to exploit the image-level labels. The adaptive bias in EM-Adapt~\cite{wsss_em} can be seen as a special case of the constrained optimization problem with just suppression and foreground constraint.

CAM~\cite{cam} significantly advanced the research on WSSS. SEC~\cite{sec} first introduces CAM to WSSS and proposes three principles for WSSS: seed, expand, and constrain, which have influenced the subsequent works. Most works in WSSS are built upon CAM and developed to address the partial activation problem (as introduced in Section~\ref{sec:intro}). We categorize those methods into four groups based on their operational level: pixel-wise, image-wise, cross-image, and external data. We will discuss each of these categories in this section.

\subsection{Pixel-Wise Methods}
In WSSS, even though we are limited to image-level labels, several methods delve into information at the pixel level. Some methods, such as ~\cite{amn,sbce,sance}, derive pixel-level supervision signals from the image-level labels and utilize them to optimize pixel-wise loss functions. Others, like \cite{psa,irn,auxseg,afa,sas}, explore the similarity among neighboring pixels. Expanding beyond individual pixels, certain approaches\cite{l2g,cpn,rpim} harness the complementary information from local patches comprised of multiple pixels. Furthermore, there are other methods that explore contrastive learning~\cite{ppc,toco}, graph convolution network~\cite{nsrom} on the pixel level.

\noindent
\subsubsection{Pixel-wise loss.} 
In WSSS with image-level class labels, we lack pixel-level labels for direct network supervision. As a result, certain methods have been developed to generate pixel-level supervision using diverse strategies. 
A straightforward strategy is to utilize the seed (or refined with dense Conditional Random Field (dCRF)~\cite{dcrf}) as noisy supervisory. AMN~\cite{amn} strives to increase the activation gap between the foreground and background regions. This ensures the resultant pseudo masks are robust to the global threshold values utilized to separate the foreground and background. To achieve this, AMN develops an activation manipulation network equipped with a per-pixel classification loss function (balanced cross-entropy loss~\cite{dsrg}), which is supervised by the confident regions within the refined seeds. SANCE~\cite{sance} trains a model to predict both object contour map and segmentation map, supervised by noisy seeds and online label simultaneously. This method employs noisy seeds as supervision for the segmentation branch and then refines the segmentation map to generate online labels to offer more accurate semantic supervision to the contour branch. Finally, they can generate more complete pseudo masks based on the segmentation map and the contour map. In Spatial-BCE~\cite{sbce}, the authors highlighted a drawback of the traditional BCE loss function: it calculates the average over the entire probability map (i.e., via global average pooling), thereby causing all pixels to be optimized in the same direction. This process reduces the discriminative ability between the foreground and background pixels. To address this issue, they proposed a spatial BCE loss function that optimizes foreground and background pixels in distinct directions. An adaptive threshold is employed to divide the foreground and background within the initial seeds.

All three approaches share a crucial commonality: the use of a threshold to discern between foreground and background. This threshold stands as a pivotal determinant, given that it classifies each pixel as either background or foreground, thereby profoundly affecting the subsequent learning trajectory. The first two methods~\cite{amn,sance} employ a fixed threshold, set as a hyper-parameter. In contrast, the latter approach~\cite{sbce} opts for a learnable threshold which can be optimized in the training process.

\noindent
\subsubsection{Pixel similarity.} 
These kinds of methods leverage the similarity between adjacent pixels to refine seeds. PSA~\cite{psa}, IRN~\cite{irn}, AuxSegNet~\cite{auxseg}, and AFA~\cite{afa} propagate the object regions in the seeds to semantically similar pixels in the neighborhood. It is achieved by the random walk~\cite{randomwalk} on a transition matrix where each element indicates an affinity score of two pixels. PSA~\cite{psa} introduces an AffinityNet to directly predict semantic affinities between adjacent pixels. IRN~\cite{irn} incorporates an inter-pixel relation network to estimate class boundary maps. Based on the assumption that a class boundary exists between pairs of pixels with different pseudo class labels, they compute affinities from the predicted class boundary maps. PSA and IRN then use affinity labels, computed by the initial CAM, as supervision. This is because CAM is often locally accurate and provides evidence to identify semantic affinities within a small image area. AuxSegNet~\cite{auxseg} integrates non-local self-attention blocks, which capture the semantic correlations of spatial positions based on the similarities between the feature vectors of any two positions. 

The propagation of CAM in those methods can effectively reduce the false negatives in the original CAM. However, one potential limitation is that the random walk on transition matrix can be time-intensive. 
Different from them, AFA~\cite{afa} and SAS~\cite{sas} leverage the pixel similarity from the inherent self-attention in the Transformer-based backbone. Specifically, Ru et~al.~\cite{afa} introduced an Affinity from Attention (AFA) module to learn semantic affinities from the multi-head self-attention (MHSA) in Transformers. They generate an initial CAM and then use it to compute pseudo affinity labels, representing pixel similarity. These pseudo affinity labels are subsequently utilized to guide the affinity prediction made by the MHSA. Kim et~al.~\cite{sas} proposed a superpixel discovery method to find the semantic-aware superpixels based on the pixel-level feature similarity produced by self-supervised vision transformer~\cite{dino}. Then the superpixels are utilized to expand the initial seed.

\noindent
\subsubsection{Local patch.} 
Moving beyond the scope of a single pixel, certain methods operate within the context of a small patch composed of a cluster of adjacent pixels.
CPN~\cite{cpn} demonstrates that the self-information of the CAM of an image is less than or equal to the sum of the self-information of the CAMs, which are obtained by complementary patch pair. They split an image into two images with complementary patch regions and used the sum of CAMs generated by the two images to mine out more foreground regions. L2G~\cite{l2g} employs a local classification network to extract attention from various randomly cropped local patches within the input image. Concurrently, it uses a global network to learn complementary attention knowledge across multiple local attention maps online. Different from CPN~\cite{cpn} and L2G~\cite{l2g}, where the patches are randomly divided, RPIM~\cite{rpim} utilizes the superpixel approach to partition the input images into different regions. It then uses an inter-region spreading module to discover the relationship between regions and merge the regions that belong to the same object into a whole semantic region.

\noindent
\subsubsection{Other pixel-wise methods.}
There are also other methods that operate at the pixel level, but they don't fit neatly into the previously mentioned categories. 
RRM~\cite{rrm} finds that the reliability of the pseudo masks is important. They proposed to only choose those conﬁdent object/background regions that are usually tiny but with high response scores on the class activation maps.
1-stage~\cite{1stage} introduces the normalized global weighted pooling that utilizes pixel-level confidence predictions for relative weighting of the corresponding classification score. It employs a stochastic gate that mixes feature representations with varying receptive field sizes to ensure neighboring pixels with similar appearances share the same label.
Fan et~al.~\cite{icd} proposed an intra-class discriminator (ICD) that is dedicated to separating the foreground and the background pixels within each image-level class. Such an intra-class discriminator is similar to a binary classifier for each image-level class, which identifies between the foreground pixels and the background pixels.
NSROM~\cite{nsrom} performs the graph-based global reasoning~\cite{gnn} on pixel-level feature maps to strengthen the classification network’s ability to capture global relations among disjoint and distant regions. This helps the network activate the object features outside the salient area.
DRS~\cite{drs} takes a unique approach by attempting to suppress the discriminative regions, thereby redirecting attention to adjacent non-discriminative regions. This is accomplished by introducing suppression controllers (which can be either learnable or non-learnable) to each layer of the CNNs, controlling the extent to which the attention is focused on discriminative regions. Specifically, any activation values that exceed a certain threshold (which can be fixed or learnable) multiplied by the maximum activation values will be suppressed to the maximum activation values. This methodology ensures that the model's focus is more evenly distributed across the whole image, rather than being concentrated solely on the most distinctive regions.
PPC~\cite{ppc} is instantiated with a unified pixel-to-prototype contrastive learning formulation, which shapes the pixel embedding space through a prototype-based metric learning methodology. The core idea is pulling pixels together to their positive prototypes and pushing them away from their negative prototypes to learn discriminative dense visual representations.
ToCo~\cite{toco} devises a Class Token Contrast (CTC) module inspired by the capability of ViT's class tokens to capture high-level semantics. CTC harnesses reliable foreground and background regions within the initial CAM to derive positive and negative local images. The class tokens of these positive/negative images are then projected and contrasted with the global class token using InfoNCE loss~\cite{infonce}, assisting in differentiating low-confidence regions within the CAM.

\subsection{Image-Wise Methods}
Image-wise methods are the most straightforward and have been the subject of numerous works. Researchers employing these methods have explored a diverse array of strategies.
A considerable number of studies delve into adversarial learning~\cite{adv_erasing,cse,aeft,ecsnet,advcam,acr}, context decoupling~\cite{conta,cda,bdm,ccam}, and consistency regularization~\cite{seam,sipe,amr}. Some tackle the problems of loss functions, introducing innovative solutions. Furthermore, several methods have emerged that focus on online attention accumulation~\cite{ooa}, uncertainty estimation~\cite{urn}, and evaluating CAM's coefficient of variation~\cite{pmm}.

\noindent
\subsubsection{Adversarial learning.} 
The first kind of method that leverages the idea of adversarial learning is adversarial erasing (AE) based methods.
Wei et~al.~\cite{adv_erasing} proposed the first AE method, which discovers a small object region and erases the mined regions from the image. Then it feeds the image to the classification network again to drive the network to discover new and complementary object regions. After multiple feed-forward processes, they can combine those object regions to achieve a more complete mask.
Kweon et~al.~\cite{cse} proposed a class-specific AE framework that randomly samples a single specific class to be erased. It generates a class-specific mask by CAM for the target classes and erases the target class from the image. Then the erased image is fed into the classifier, and the gradients back-propagated from the class-specific erasing loss guide the network to generate more precise CAMs. Ideally, this process ensures that the erased image removes all pixels belonging to the target class while retaining pixels from other classes intact.

Although AE methods expand the CAM by erasing the most discriminative regions, they often encounter high computation cost problems due to the multiple feed-forward process and the over-expansion problem due to the lack of guidance on when to stop the erasing process. To address this, the AEFT method ~\cite{aeft} reformulates the AE methods as a form of triplet learning. Specifically, it designates the original image as an anchor image, the masked high-confidence regions of the CAM on the anchor image as a positive image, and another image (which shares no class overlap with the anchor image) as a negative image. It aims to minimize the distance between the anchor and the positive image in the feature space while simultaneously maximizing the distance between the anchor and the negative image. As a result, when the CAMs are over-expanded, the embedding from the low-confidence region includes less information about the objects in the image, making it challenging for the network to differentiate this less informative embedding from the negative embedding. Consequently, the expansion of CAMs is intuitively suppressed.
ECS-Net~\cite{ecsnet} investigates a way to provide additional supervision for the classification network by utilizing predictions of erased images. It first erases high-response regions from images and generates new CAMs of those erased images. Then, it samples reliable pixels from the new CAM and applies their segmentation predictions as semantic labels to train the corresponding original CAM. Instead of erasing multiple times, ECS-Net only needs to erase once, avoiding introducing excessive noise.

Instead of directly erasing the mined regions from the image, AdvCAM~\cite{advcam} perturbs the image along pixel gradients which increases the classification score of the target class. The result is that non-discriminative regions, which are nevertheless relevant to that class, gradually become involved in the classification. However, a notable drawback of AdvCAM is the computation of these gradients, which is computationally intensive and significantly slows down the process.

Unlike all those methods, kweon~et~al.~\cite{acr} presented a framework that utilizes adversarial learning between a classifier and an image reconstructor. This method is inspired by the notion that no individual segment should be able to infer color or texture information from other segments if semantic segmentation is perfectly achieved. They introduced an image reconstruction task that aims to reconstruct one image segment from the remaining segments. The classifier is trained not only to classify the image but also to generate CAM that accurately segment the image, while contending with the reconstructor. In the end, the quality of the CAMs is enhanced by jointly training the classifier and the reconstructor in an adversarial manner.

\noindent
\subsubsection{Context decoupling.} 
Some methods attempt to decouple the object from its surrounding context. 
For instance, Zhang et~al.~\cite{conta} proposed a structural causal model (CONTA) to analyze the causalities among images, their contexts, and class labels. Based on this, they developed a context adjustment method that eliminates confounding bias in the classification model, resulting in improved CAM. 
CDA~\cite{cda} is a context decoupling augmentation technique that modifies the inherent context in which objects appear, thereby encouraging the network to remove reliance on the correlation between object instances and contextual information. Specifically, in the first stage, it uses the off-the-shelf WSSS methods to obtain basic object instances with high-quality segmentation. In the second, these object instances are randomly embedded into raw images to form the new input images. These images then undergo online data augmentation training in a pairwise manner with the original input images. 
Unlike CDA, which relies on pre-existing WSSS methods to separate background and foreground, BDM~\cite{bdm} utilizes saliency maps to generate a binary mask, cropping out images containing only the background or foreground for a given image. It subsequently applies consistency regularization to the CAMs derived from object instances seen in various scenes, thereby providing self-supervision for network training.
Different from CDA and BDM, which apply a mask on the original image to decouple the foreground and background. CCAM~\cite{ccam} generates a class-agnostic activation map and disentangles image representation into the foreground and background representations. These disentangled representations are then used to create positive pairs (either foreground-foreground or background-background representations) and negative pairs (foreground-background representations) across a group of images. Finally, using these constructed pairs, a contrastive loss function is applied to encourage the network to effectively separate foreground and background.

\noindent
\subsubsection{Consistency regularization.}
These methods leverage consistency regularization to guide network learning.
SEAM~\cite{seam} employs consistency regularization on predicted CAMs from various transformed images to provide self-supervision for network learning. 
SIPE~\cite{sipe} ensures the consistency between CAM and the proposed Image-Specific Class Activation Map (IS-CAM), which is derived from image-specific prototypes. Specifically, they defined foreground and background prototypes as the centroid of seed regions in hierarchical feature space. This process performs class-wise compression on the seed pixels, achieving $K$ (number of classes) foreground prototypes and one background prototype. Then the IS-CAM can be calculated by replacing classifier weights $\mathbf{w}_{k}$ in Eq.~\ref{equation:cam} with image-specific prototypes.
Qin et~al.~\cite{amr} proposed an activation modulation and re-calibration scheme that leverages a spotlight branch and a compensation branch to provide complementary and task-oriented CAMs. The spotlight branch denotes the fundamental classification network, while the compensation branch contains an attention modulation module to rearrange the distribution of feature importance from the channel-spatial sequential perspective to dig out the important but easily ignored regions. A consistency loss is employed between the CAMs produced by the two branches.

The consistency regularization term is a versatile component that can be integrated into various network designs, provided there are coherent features or CAMs available. Typically presented as an auxiliary loss function, it enhances the model's robustness without adding significant computational demands. This makes it an attractive add-on that ensures consistent feature representations across different stages of the model.

\noindent
\subsubsection{Loss function.} 
Some methods investigate the existing problems associated with the BCE loss function used in classification models and propose new loss functions to mitigate these issues. For instance, Lee~et~al.~\cite{rib} highlighted that the final layer of a deep neural network, activated by sigmoid or softmax activation functions, often leads to an information bottleneck. To counter this, they proposed a novel loss function that eliminates the final non-linear activation function in the classification model while also introducing a new pooling method that further promotes the transmission of information from non-discriminative regions to the classification task.
Similarly, Chen~et~al.~\cite{recam} identified a problem with the widespread use of BCE loss --- it fails to enforce class-exclusive learning, often leading to confusion between similar classes. They proved the superiority of softmax cross-entropy (SCE) loss and suggested integrating SCE into the BCE-based model to reactivate the classification model. Specifically, they masked the class-specific CAM on the feature maps and applied SCE loss on the masked feature maps, thereby facilitating better class distinction.

Both methods identify shortcomings in the prevailing BCE loss, specifically addressing the information bottleneck problem~\cite{rib} and the class-exclusive learning problem~\cite{recam}. They each introduce new loss functions to tackle these specific issues. Recognizing the pivotal role that the loss function plays in network learning, both methods contribute substantially to performance enhancements. 

\noindent
\subsubsection{Other image-wise methods.}
There are also other methods that operate at the image level, but they don't fit neatly into the previously mentioned categories.
Jiang~et~al.~\cite{ooa} proposed an online attention accumulation (OAA) strategy that maintains a cumulative attention map for each target category in each training image to accumulate the discovered different object parts. So that the integral object regions can be gradually promoted as the training goes.
EDAM~\cite{edam} masks the class-specific CAM on the feature maps and learns separate classifiers for each class. As the region of background and irrelevant foreground objects are removed in the class-specific feature map, to some extent, the performance of classification can be improved.
PMM~\cite{pmm} computes the coefficient of variation for each channel of CAMs and then refines CAMs via exponential functions with the coefficient of variation as the control coefficient. This operation smooths the CAMs and could alleviate the partial response problem introduced by the classification pipeline.
URN~\cite{urn} simulates noisy variations of the prediction by scaling the prediction map multiple times for uncertainty estimation. The uncertainty is then used to weigh the segmentation loss to mitigate noisy supervision signals.
ESOL~\cite{esol} employs an Expansion and Shrinkage scheme based on the offset learning in the deformable convolution~\cite{deformable_conv}, to sequentially improve the recall and precision of the located object in the two respective stages. The Expansion stage aims to recover the entire object as much as possible by sampling the exterior object regions beyond the most discriminative ones. It can improve the recall of the located object regions. The Shrinkage stage excludes the false positive regions and thus further enhances the precision of the located object regions.
MCTFormer~\cite{mct} uses multiple class tokens in a transformer-based network to learn the interactions between these class tokens and the patch tokens. This allows the model to learn class-specific activation maps from the class-to-patch attention of various class tokens.
TSCD~\cite{tscd} proposes self-correspondence distillation (SCD) to enhance the network feature learning process. It utilizes the CAM feature correspondence as the distillation target for segmentation prediction features. SCD aligns the segmentation feature correspondence with the network’s own CAM feature correspondence. This mechanism can help the network obtain comprehensive image semantic information and thus generate high-quality CAM.

\subsection{Cross-Image Methods}
Moving beyond a single image, certain connections often exist between different images within a dataset. Some methods explore these connections, whether they occur pair-wise~\cite{mcis,cian,mbmnet}, group-wise~\cite{hgnn,group,ccam}, or even on a dataset-wise scale~\cite{sce,rca,lpcam}.

\noindent
\subsubsection{Pair-wise methods.} 
Some methods focus on capturing pairwise relationships between images. For instance, MCIS~\cite{mcis} employs two neural co-attentions in its classifier to capture complementary semantic similarities and differences across images. Given a pair of training images, one co-attention forces the classifier to recognize the common semantics from co-attentive objects, while the other drives the classifier to identify the unique semantics from the rest, uncommon objects. This dual attention approach helps the classifier discover more object patterns and better ground semantics in image regions.
Similarly, Fan~et~al.~\cite{cian} proposed an end-to-end cross-image affinity module designed to gather supplementary information from related images. Specifically, it builds pixel-level affinities across different images, allowing incomplete regions to glean additional information from other images. This approach results in more comprehensive object region estimations and mitigates ambiguity.
MBMNet~\cite{mbmnet} utilizes a parameter-shared siamese encoder to encode the representations of paired images and models their feature representations with a bipartite graph. They find the maximum bipartite matching (MBM) between the graph nodes to determine relevant feature points in two images, which are then used to enhance the corresponding representations.
CTI~\cite{cti} is developed on the class tokens of the ViT~\cite{vit} backbone. It tries to guide class tokens to possess class-specific distinct characteristics and global-local consistency by infusing the class tokens from paired images. Specifically, it infuses class-specific characteristics from other images that share at least one same class and global-local consistency from the same image with different transformations.

\noindent
\subsubsection{Group-wise methods.} 
Some methods attempt to model more complex relationships within a group of images. For instance, Group-WSSS~\cite{group} explicitly models semantic dependencies in a group of images to estimate more reliable pseudo masks. Specifically, they formulate the task within a graph neural network (GNN)~\cite{gnn}, which operates on a group of images and explores their semantic relations for more effective representation learning.
Additionally, Zhang~et~al.~\cite{hgnn} introduced a heterogeneous graph neural network (HGNN) to model the heterogeneity of multi-granular semantics within a set of input images. The HGNN comprises two types of sub-graphs: an external graph and an internal graph. The external graph characterizes the relationships across different images, aiming to mine inter-image contexts. The internal graph, which is constructed for each image individually, is used to mine inter-class semantic dependencies within each individual image. Through heterogeneous graph learning, the network can develop a comprehensive understanding of object patterns, leading to more accurate semantic concept grounding.

In these methods, it is important to determine the number of images in a group. There's a delicate balance between capturing meaningful semantic relationships and introducing noise by grouping images. Both approaches utilize groups of $4$ images. As the number of images increases, the benefits from additional semantic cues plateau, while the noise introduced can lead to a decline in performance.

\noindent
\subsubsection{Dataset-wise methods.} 
Beyond a group of images, some methods delve into the semantic connections present in the entire dataset.
Chang~et~al.~\cite{sce} introduced a self-supervised task leveraging sub-category information. To be more specific, for each class, they perform clustering on all local features (features at each spatial pixel position in the feature maps) within that class to generate pseudo sub-category labels. They then construct a sub-category objective that assigns the network a more challenging classification task.
Similarly, LPCAM~\cite{lpcam} also performs clustering on local features. However, instead of creating a sub-category objective, LPCAM utilizes the clustering centers, also known as local prototypes, as a non-biased classifier to compute the CAM. Since these local prototypes contain rich local semantics like the ``head'', ``leg'', and ``body'' of a ``sheep'', they are able to solve the problem that the weight of the classifier (which is used to compute the CAM) in the classification model only captures the most discriminative features of objects.
Zhou~et~al.~\cite{rca} proposed the Regional Semantic Contrast and Aggregation (RCA) method for dataset-level relation learning. RCA uses a regional memory bank to store a wide array of object patterns that appear in the training data, offering strong support for exploring the dataset-level semantic structure. More specifically, the semantic contrast pushes the network to bring the embedding closer to the memory embedding of the same category while pushing away those of different categories. This contrastive property complements the classification objective for each individual image, thereby improving object representation learning. On the other hand, semantic aggregation enables the model to gather dataset-level contextual knowledge, resulting in more meaningful object representations. This is achieved through a non-parametric attention module which summarizes memory representations for each image independently.

Compared to the pair-wise and group-wise methods, the dataset-wise methods can leverage more information from the whole dataset. These methods explore the object patterns on the whole dataset. These patterns are then captured and stored using mechanisms like ``sub-category labels''~\cite{sce}, ``local prototypes''~\cite{lpcam}, or ``memory banks''~\cite{rca}, which can be later leveraged for model learning or CAM generation. These object patterns can effectively improve the CAM quality. Similar to group-wise methods, one challenge is the balance of meaningful semantic cues and potential noise. Thus, it always needs some operations, like the selection of prototypes in LPCAM~\cite{lpcam}, to preserve useful object patterns.

\subsection{Methods with External Data}
In addition to leveraging information within the dataset, some methods further employ external data resources to improve the model. These external resources can provide additional, diverse, and complementary information not present in the original dataset, helping to improve the model's overall performance.

\noindent
\subsubsection{Saliency map.} 
Saliency detection methods~\cite{saliency,saliency1,saliency2} generate saliency maps that distinguish between the foreground and the background in an image. Many WSSS methods~\cite{edam,ficklenet,ooa,dsrg,mcis,rca} exploit saliency maps as a post-processing step to refine the initial CAMs. Beyond their usage in post-processing, some methods employ saliency maps to aid model learning.
For instance, both SSNet~\cite{ssnet} and EPS~\cite{eps} directly model the connection between saliency detection and WSSS. They achieve this by jointly minimizing the classification loss and saliency loss. The saliency loss is defined as the pixel-wise difference between the actual saliency map and the estimated saliency map. Both methods use a classifier to predict localization maps (or segmentation masks) from backbone features. However, their approaches to estimating the saliency map vary. SSNet~\cite{ssnet} employs the saliency aggregation module to predict the saliency score of each category and then aggregates the segmentation masks of all categories into a saliency map according to their saliency scores. EPS~\cite{eps} estimates the saliency maps by computing a weighted sum of foreground and background maps. By introducing the background map into saliency maps, EPS can mitigate the issue of co-occurring pixels of non-target objects.

These methods leverage a saliency detection model to generate the saliency maps, eliminating the need for extra human effort to annotate. However, there's no guarantee that the saliency maps will be perfect. Consequently, inaccuracies in the saliency maps can impact the overall performance of the method.

\noindent
\subsubsection{Out-of-distribution data.}
Lee et~al.~\cite{ood} proposed to use Out-of-Distribution (OoD) data to address the issue of spurious correlation between foreground and background cues (e.g., ``train'' and ``rail''). They collected their candidate OoD data, which do not include any foreground classes of interest, from another vision dataset OpenImages~\cite{openimages}. Taking the class ``train'' as an example, they initially select images in which the classification model's predicted probability for ``train'' exceeded 0.5. They then manually filter out images that contained a ``train''. The remaining images, which do not contain a "train" but have a high predicted probability for the class "train" in the classification model, can be used as out-of-distribution data. They assign out-of-distribution data with zero-vector labels (zero for all classes) and apply the common binary cross-entropy (BCE) loss for both in-distribution and out-of-distribution samples. The OoD data helps the model distinguish between in-distribution and out-of-distribution samples, thereby reducing false positive predictions in CAM. 

Utilizing OoD data can significantly enhance the model's ability to distinguish between objects and co-occurring background cues.  However, the annotation of OoD data requires additional human effort.
\section{Foundation Models}
\label{sec:foundation}
\subsection{Contrastive Language-Image Pre-Training (CLIP)}
Contrastive Language-Image Pre-Training (CLIP)~\cite{clip} is designed to efficiently learn visual concepts from natural language supervision. The main innovation of CLIP is its use of a contrastive objective function, which is based on the principle that semantically similar inputs should be mapped to nearby points in the feature space, while semantically dissimilar inputs should be mapped to distant points. Specifically, CLIP is trained on a large dataset of image-text pairs using contrastive loss. The loss encourages the model to learn to map similar image and text representations close together in a joint feature space while pushing dissimilar representations apart. CLIP has emerged as a powerful tool due to its ability to associate much wider visual concepts in the image with their text labels in an open-world setting. Thus, it has been applied in several semantic segmentation tasks like zero-shot semantic segmentation~\cite{zegformer,zsseg,maskclip,zegclip}, open-vocabulary semantic segmentation~\cite{ovs_clip1,ovs_clip2,unpair_seg}, and weakly-supervised semantic segmentation~\cite{clims,clipes,weclip}.

CLIP has been applied to zero-shot semantic segmentation via a two-stage scheme~\cite{zegformer,zsseg}. The general idea is to first generate class-agnostic region proposals using a pretrained mask proposal generator and then feed the cropped proposal regions to CLIP to utilize its image-level zero-shot classification capability. Although CLIP’s zero-shot ability remains at the image level, the computational cost will inevitably increase due to the classification of each proposal. 
The one-stage solution can solve this issue by directly extending CLIP’s zero-shot prediction capability from image to pixel level. MaskCLIP+~\cite{maskclip} utilizes the CLIP text encoder to produce text embeddings of target classes. These text embeddings are then applied as classifiers on the image embeddings to generate pseudo masks. Subsequently, a segmentation network, such as DeepLabv2\cite{deeplabv2}, is trained using the generated pseudo masks. ZegCLIP~\cite{zegclip} uses prompt tuning on the CLIP image encoder and a lightweight decoder to match the text embedding with the patch-level image embeddings extracted from CLIP. Compared to MaskCLIP+, ZegCLIP can infer both seen and unseen classes, whereas MaskCLIP+ is limited to seen classes.

Several works have harnessed the potential of CLIP in WSSS. CLIMS~\cite{clims} utilizes the CLIP model as a text-driven evaluator. Specifically, it employs a CNN to generate initial CAMs and applies the CAMs (or reversed CAMs) on the image as masks to identify the object (or background) regions. It then leverages object text prompts (e.g., ``a photo of a train'') and class-related background text prompts (e.g., ``a photo of railroad'' for class ``train'') to compute matching losses with the masked object and background regions, respectively. These losses work to ensure both the correctness and completeness of the initial CAMs. Different from CLIMS~\cite{clims}, CLIP-ES~\cite{clipes} investigates the ability of CLIP to localize different categories through only image-level labels without any additional training. To efficiently generate high-quality segmentation masks from CLIP, they propose a framework with special designs for CLIP. They introduce the softmax function into GradCAM~\cite{gradcam} and define a class-related background set to enforce mutual exclusivity among categories, thus suppressing the confusion caused by non-target classes and backgrounds. Meanwhile, to take full advantage of CLIP, they re-explore text inputs under the WSSS setting and customize two text-driven strategies: sharpness-based prompt selection and synonym fusion. 
Both CLIMS and CLIP-ES only use the CLIP model to improve CAM for better pseudo labels. Different from them, WeCLIP~\cite{weclip}  explores the potential of the CLIP model to be directly used as the backbone to extract strong semantic features for segmentation prediction. They designed a frozen CLIP feature decoder based on the transformer architecture to interpret the frozen features for semantic prediction.

\subsection{Segment Anything Model (SAM)}
\begin{figure}[ht]
\centering
\includegraphics[width=0.95\linewidth]{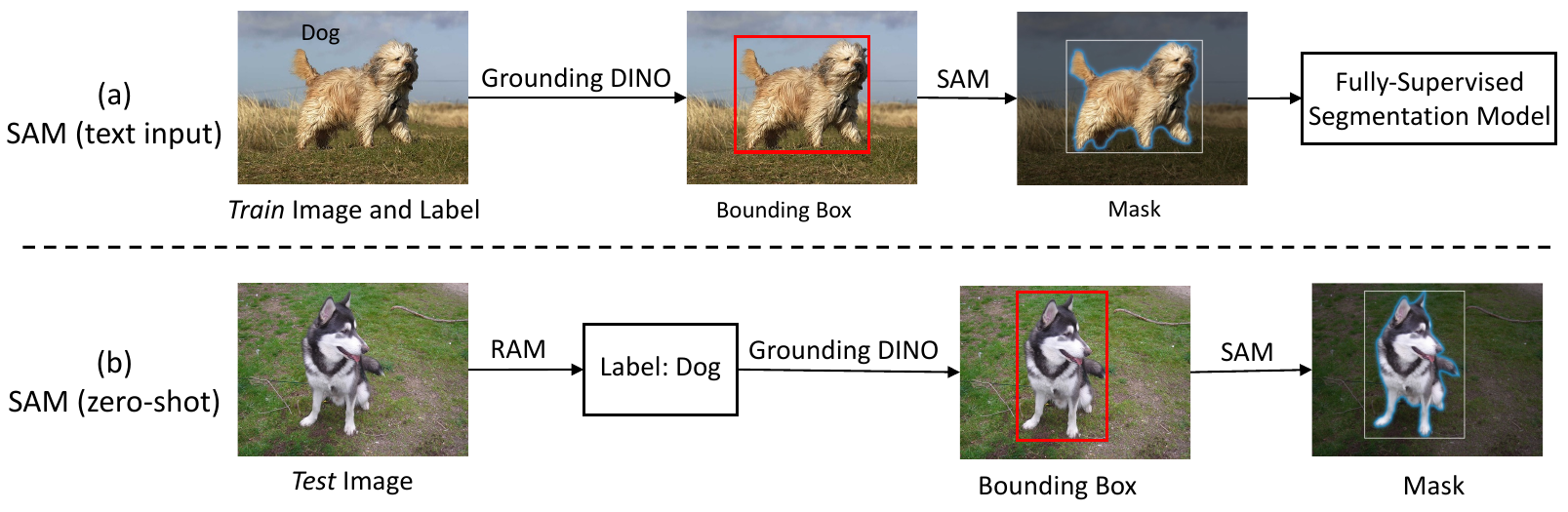}
\caption{The pipeline of applying SAM in WSSS. All models except the fully-supervised segmentation model are kept frozen.}
\Description{
This figure illustrates the pipeline for applying SAM in weakly supervised semantic segmentation (WSSS). Panel (a) shows the process of using SAM with text input prompts, where Grounding DINO generates bounding boxes and SAM produces masks from labeled training images. These masks are subsequently utilized to train the fully-supervised segmentation model. Panel (b) represents the zero-shot setting, where the RAM model predicts object labels for test images. These labels are processed by Grounding DINO to produce bounding boxes, and SAM generates the final masks. Throughout both stages, all models except the fully-supervised segmentation model remain frozen.
}
\label{fig:pipeline_sam}
\end{figure}

The Segment Anything Model (SAM)~\cite{sam} is a recent image segmentation model exhibiting superior performance across various segmentation tasks. Different from the traditional semantic segmentation model~\cite{deeplabv3+,setr,seit} where the input is an image and the output is a mask, SAM introduces a new promptable segmentation task that supports various types of prompts, such as points, bounding boxes, and textual descriptions. It leverages a Transformer model~\cite{vit} trained on the extensive SA-1B dataset (comprising over 1 billion masks derived from 11 million images), which gives it the ability to handle a wide range of scenes and objects. SAM is remarkable for its capability to interpret diverse prompts and successively generate various object masks. We investigate two settings to apply SAM in WSSS: text input (as the \texttt{train} images have class labels in WSSS) and zero-shot (as there is no class label for the \texttt{val} and \texttt{test} images in WSSS). The pipeline of the two settings is shown in Fig.~\ref{fig:pipeline_sam}.

\subsubsection{SAM (text input)}
We follow the general pipeline in traditional methods (as shown in Fig.~\ref{fig:pipeline}) to generate pseudo masks first and then train the fully supervised semantic segmentation model (DeepLabV2~\cite{deeplabv2}). To generate pseudo masks, we consider feeding image and text prompts (class labels) as inputs to SAM. However, the text prompt functionality of SAM is not currently open-sourced. To circumvent this limitation, as shown in Fig.~\ref{fig:pipeline_sam} (a), we utilize Grounded-SAM~\footnote{\href{https://github.com/IDEA-Research/Grounded-Segment-Anything}{https://github.com/IDEA-Research/Grounded-Segment-Anything}} in our experiments for WSSS. Grounded-SAM is a hybrid of Grounding DINO~\cite{groundingdino} and SAM~\cite{sam}, enabling the grounding and segmentation of objects via text inputs. In particular, Grounding DINO can generate grounded bounding boxes using text prompts. Following this step, we feed these grounded bounding boxes into SAM to produce corresponding segmentation masks. The pipeline of SAM (text input) is summarized as follows:
\begin{itemize}
    \item Step~1: Feed the \texttt{train} images and image-level class labels to Grounding~DINO~\cite{groundingdino} to generate bounding boxes.
    \item Step~2: Feed the \texttt{train} images and generated bounding boxes to SAM~\cite{sam} to generate masks.
    \item Step~3: Train the fully-supervised segmentation model (e.g. DeepLabV2~\cite{deeplabv2}) using the \texttt{train} images and generated masks.
    \item Step~4: Feed the \texttt{test} images to segmentation model to generate masks.
\end{itemize}

\subsubsection{SAM (zero-shot)}
Different from the general pipeline (as shown in Fig.~\ref{fig:pipeline}) used in traditional WSSS, we also endeavor to directly assess SAM's performance on the \texttt{val} and \texttt{test} images, which lack any text inputs. As illustrated in Fig.~\ref{fig:pipeline_sam} (b), we utilize the image tagging model, Recognize Anything Model (RAM)~\cite{ram}, to identify the objects within the image first. RAM is a strong foundation model designed for image tagging. It demonstrates the strong zero-shot ability to recognize any category with high accuracy, surpassing the performance of both fully supervised models and existing generalist approaches like CLIP~\cite{clip} and BLIP~\cite{blip}. With the recognized category label, we adhere to the similar pipeline as SAM (text input). The pipeline of SAM (zero-shot) is summarized as follows:

\begin{itemize}
    \item Step~1: Feed the \texttt{test} images to RAM~\cite{ram} to generate tags.
    \item Step~2: Feed the \texttt{test} images and generated tags to Grounding~DINO~\cite{groundingdino} to generate bounding boxes.
    \item Step~3: Feed the \texttt{test} images and generated bounding boxes to SAM~\cite{sam} to generate masks.
\end{itemize}

Compared to SAM (text input), SAM (zero-shot) incorporates an additional model (SAM) to identify objects within the image. Notably, the SAM (zero-shot) method eliminates the need for \texttt{train} images and eliminates the need to train a fully supervised segmentation model in SAM (text input). This is because SAM (zero-shot) can predict masks even in the absence of class labels.
\section{Methodological Comparison}
\label{sec:comparison}
In this section, we provide a comprehensive comparison between the traditional models introduced in Section~\ref{sec:tradition} and the application of foundation models introduced in Section~\ref{sec:foundation}. We also offer insights into the potential and challenges of deploying foundational models in WSSS.

\subsection{Datasets and Implementation Details}
\subsubsection{Datasets}
There are two benchmark datasets in WSSS: PASCAL VOC 2012~\cite{voc} and MS~COCO 2014~\cite{mscoco}. PASCAL VOC 2012 contains $20$ foreground object categories and $1$ background category with $1,464$ \texttt{train} images, $1,449$ \texttt{val} images, and $1,456$ \texttt{test} images. All works use the enlarged training set with $10,582$ training images provided by SBD~\cite{voc_aug}. MS~COCO 2014 dataset consists of $80$ foreground categories and $1$ background category, with $82,783$ and $40,504$ images in \texttt{train} and \texttt{val} sets, respectively.

\subsubsection{Metrics}
There are two evaluation steps for WSSS --- the pseudo mask quality and the semantic segmentation performance. 
The pseudo mask quality is evaluated by the mean Intersection-over-Union (mIoU) of the generated pseudo masks and the corresponding ground truth masks on the \texttt{train} images.
In terms of the semantic segmentation performance, we evaluate the mIoU between the predicted masks and the corresponding ground truth masks on both the \texttt{val} and \texttt{test} images.

\subsubsection{Implementation details}

\noindent
\textbf{Grounded-SAM model.}
In the text input setting, we load the default pretrained Grounded-SAM model (includes Swin-T~\cite{swin} for Grounding DINO~\cite{groundingdino} and ViT-H~\cite{vit} for SAM~\cite{sam}). The Grounding DINO~\cite{groundingdino} is an open-set detector that uses Swin-T~\cite{swin} as the image encoder and BERT~\cite{bert} as the text encoder. It is trained on Objects365~\cite{objects365}, GoldG~\cite{goldg}, and Cap4M~\cite{glip}. It is trained using the AdamW~\cite{adamw} optimizer and learning rate $1e{-5}$ for image and text backbone and $1e{-4}$ for other parts. The model is trained on $64$ NVIDIA A100 GPUs with a total batch size of $64$. In the inference, the box threshold in Grounding DINO~\cite{groundingdino} is set to $0.3$, and the text threshold is set to $0.25$. The SAM is initialized from an MAE~\cite{mae} pre-trained ViT-H~\cite{vit}. It is trained using the AdamW~\cite{adamw} optimizer ($\beta_1 = 0.9, \beta_2 = 0.999$) and a linear learning rate warmup for 250 iterations and a step-wise learning rate decay schedule. The initial learning rate, after warmup, is $8e{-4}$. They train it for 90k iterations (about 2 SA-1B~\cite{sam} epochs) and decrease the learning rate by a factor of $10$ at $60k$ iterations and again at $86,666$ iterations. The batch size is set to 256 images. To regularize SAM, they set weight decay to $0.1$ and apply drop path with a rate of $0.4$. They use a layer-wise learning rate decay of $0.8$. No data augmentation is applied. SAM is trained on $256$ NVIDIA A100 GPUs for $68$ hours. The inference speed of the Grounded-SAM model is about $2$ images per second on an NVIDIA L40 GPU.

\noindent
\textbf{RAM model.}
In the zero-shot setting, we load the pre-trained RAM-14M~\cite{ram} model based on Swin-L~\cite{swin}. The model is trained on $14$M  images from various datasets, including COCO~\cite{mscoco}, Visual Genome~\cite{visual_genome}, Conceptual Captions~\cite{conceptual_12m}, and SBU Captions~\cite{sbu_captions}. It is trained on $8$ NVIDIA A100 GPUs for $3$ days. The inference speed of the combination of RAM and Grounded-SAM model is about $1.7$ images per second on an NVIDIA L40 GPU. To align the tags from the RAM with the class labels in the VOC datasets, we establish a mapping strategy due to the different terminologies used by the models and datasets. Specifically, we consider ``couch'' from RAM to correspond with ``sofa'' in the VOC dataset, ``plane'' with ``aeroplane'', ``plant'' with ``potted plant'', and ``monitor'',  ``screen'', and ``television'' with ``TV monitor''. Additionally, we map ``person'', ``man'', ``woman'', ``boy'', ``girl'', ``child'', and ``baby'' tags from RAM to the ``person'' class in VOC. This strategy ensures comparable results between SAM and the ground truth annotations of the VOC dataset. On the MS~COCO dataset, we adopt a similar strategy for aligning the RAM tags with the class labels. Due to the larger and more diverse range of categories in MS~COCO, please refer to our provided code.~\footnote{\href{https://github.com/zhaozhengChen/SAM_WSSS}{https://github.com/zhaozhengChen/SAM\_WSSS}}

\noindent
\textbf{DeepLabV2 model.}
For the DeepLabV2~\cite{deeplabv2} in the step of fully supervised semantic segmentation, we train the DeepLabV2~\cite{deeplabv2} with ImageNet~\cite{imagenet} pretrained ResNet-101~\cite{resnet}. Following~\cite{irn,rib,recam}, we crop each training image to the size of 321×321. We train the model for 20k and 100k iterations on VOC and MS COCO datasets, respectively, with the respective batch size of 5 and 10. The learning rate is set as 2.5e-4 and weight decay as 5e-4. Horizontal flipping and random crops are used for data augmentation. The training time on $2$ NVIDIA L40 GPUs is about $3$ and $24$ hours for VOC and MS COCO datasets, respectively.

\setlength{\tabcolsep}{2mm}{
\renewcommand\arraystretch{0.98}
\begin{table*}[p]
  \caption{Comparison of the WSSS methods in terms of the pseudo mask mIou(\%) and the segmentation mIoU results (\%) using DeepLabV2~\cite{deeplabv2} on VOC~\cite{voc} and MS~COCO~\cite{mscoco} dataset. ``P.M.'' denotes pseudo mask, ``RN'' denotes ``ResNet'', and ``WRN'' denotes ``WideResNet''.}
  \label{table:results}
  \centering
  \setcounter{rown}{0}
    \scalebox{0.65}{
  \begin{tabular}{llccccccccccc}
    \toprule
    &\multirow{2}*{Methods} & \multirow{2}*{Venue} & \multirow{2}*{Backbone} & \multirow{2}*{Saliency} &  \multicolumn{2}{c}{Segmentation} & \multicolumn{3}{c}{VOC} & \multicolumn{2}{c}{COCO} &\multirow{2}*{Row No.}\\
    \cmidrule(r){6-7}\cmidrule(r){8-10}\cmidrule(r){11-12}
    &&&&&Backbone&Pretrain&P.M.&Val&Test&P.M.&Val&\\
    \hline
    \multirow{19}*{\rotatebox{90}{Pixel-Level}} 
    & SANCE~\cite{sance}      & CVPR 22       & RN-101      &           & RN-101    & ImageNet  & -     & 70.9  & 72.2  & -     & 44.7  &\rownum  \\
    & AMN~\cite{amn}          & CVPR 22       & RN-50       &           & RN-101    & ImageNet  & 72.2  & 69.5  & 69.6  & 46.7  & 44.7  &\rownum  \\
    & Spatial-BCE~\cite{sbce} & ECCV 22       & WRN-38      &           & RN-101    & ImageNet  & 70.4  & 70.0  & 71.3  & -     & 35.2  &\rownum  \\
    \cmidrule(r){2-13}
    & PSA~\cite{psa}          & CVPR 18       & WRN-38      &           & WRN-38    & ImageNet  & 59.7  & 61.7  & 63.8  & -     & -     &\rownum  \\
    & IRN~\cite{irn}          & CVPR 19       & RN-50       &           & RN-101    & ImageNet  & 66.5  & 63.5  & 64.8  & 42.4  & 42.0  &\rownum  \\
    & AuxSegNet~\cite{auxseg} & ICCV 21       & WRN-38      &\checkmark & WRN-38    & ImageNet  & -     & 69.0  & 68.6  & -     & 33.9  &\rownum  \\ 
    & AFA~\cite{afa}          & CVPR 22       & MiT-B1      &           & MiT-B1    & ImageNet  & 68.7  & 66.0  & 66.3  & -     & 38.9  &\rownum  \\
    & SAS~\cite{sas}          & AAAI 23       & ViT-B/8     &           & RN-101    & ImageNet  & -     & 69.5  & 70.1  & -     & 44.8  &\rownum  \\
    \cmidrule(r){2-13}
    & CPN~\cite{cpn}          & ICCV 21       & WRN-38      &           & WRN-38    & ImageNet  & -     & 67.8  & 68.5  & -     & -     &\rownum  \\
    & L2G~\cite{l2g}          & CVPR 22       & WRN-38      &\checkmark & RN-101    & COCO      & 70.3  & 72.1  & 71.7  & -     & 44.2  &\rownum  \\
    & RPIM~\cite{rpim}        & ACMMM 22      & WRN-38      &\checkmark & RN-101    & COCO      & 69.3  & 71.4  & 71.4  & -     & -     &\rownum  \\ 
    \cmidrule(r){2-13}
    & RRM~\cite{rrm}          & AAAI 20       & WRN-38      &           & WRN-38    & ImageNet  & -     & 62.6  & 62.9  & -     & -     &\rownum \\ 
    & 1-stage~\cite{1stage}   & CVPR 20       & WRN-38      &           & WRN-38    & ImageNet  & 66.9  & 62.7  & 64.3  & -     & -     &\rownum \\
    & ICD~\cite{icd}          & CVPR 20       & VGG-16      &           & RN-101    & ImageNet  & -     & 64.1  & 64.3  & -     & -     &\rownum  \\
    & NSROM~\cite{nsrom}      & CVPR 21       & RN-50       &           & RN-101    & ImageNet  &       & 68.3  & 68.5  & -     & -     &\rownum  \\
    & DRS~\cite{drs}          & AAAI 21       & VGG-16      &\checkmark & RN-101    & ImageNet  & -     & 71.2  & 71.4  & -     & -     &\rownum  \\
    & PPC~\cite{ppc}+SEAM     & CVPR 22       & WRN-38      &           & WRN-38    & ImageNet  & 69.2  & 67.7  & 67.4  & -     & -     &\rownum  \\
    & PPC~\cite{ppc}+EPS      & CVPR 22       & WRN-38      &\checkmark & RN-101    & ImageNet  & 73.3  & 72.6  & 73.6  & -     & -     &\rownum  \\
    & ToCo~\cite{toco}        & CVPR 23       & ViT-B/16    &           & ViT-B/16  & ImageNet  & 72.2  & 69.8  & 70.5  & -     & 41.3  &\rownum  \\
    \hline
    \multirow{23}*{\rotatebox{90}{Image-Level}} 
    & AE~\cite{adv_erasing}   & CVPR 17       & VGG-16      &           & VGG-16    & ImageNet  & -     & 55.5  & 55.7  & -     & -     &\rownum  \\
    & CSE~\cite{cse}          & ICCV 21       & WRN-38      &           & WRN-38    & ImageNet  & -     & 68.4  & 68.2  & -     & 36.4  &\rownum  \\
    & ECS-Net~\cite{ecsnet}   & ICCV 21       & WRN-38      &           & WRN-38    & ImageNet  & -     & 66.6  & 67.6  & -     & -     &\rownum  \\
    & AdvCAM~\cite{advcam}    & CVPR 21       & RN-50       &           & RN-101    & ImageNet  & 70.5  & 68.1  & 68.0  & -     & -     &\rownum  \\
    & AEFT~\cite{aeft}        & ECCV 22       & WRN-38      &           & WRN-38    & ImageNet  & 71.0  & 70.9  & 71.7  & -     & 44.8  &\rownum  \\ 
    & ACR~\cite{acr}          & CVPR 23       & WRN-38      &           & WRN-38    & ImageNet  & 72.3  & 71.9  & 71.9  & -     & 45.3  &\rownum  \\
    \cmidrule(r){2-13}
    & CONTA~\cite{conta}      & NeurIPS 20    & RN-50       &           & RN-101    & ImageNet  & 67.9  & 65.3  & 66.1  & -     & 33.4  &\rownum  \\
    & CDA~\cite{cda}          & ICCV 21       & RN-50       &           & RN-101    & ImageNet  & 67.7  & 65.8  & 66.4  & -     & 33.7  &\rownum  \\
    & BDM~\cite{bdm}          & ACMMM 22      & WRN-38      &\checkmark & RN-101    & ImageNet  & 72.3  & 71.0  & 71.0  & -     & 36.7  &\rownum  \\ 
    & CCAM~\cite{ccam}        & CVPR 22       & WRN-38      &           & WRN-38    & ImageNet  & 65.5  & -     & -     & -     & -     &\rownum  \\
    \cmidrule(r){2-13}
    & SEAM~\cite{seam}        & CVPR 20       & WRN-38      &           & WRN-38    & ImageNet  & -     & 64.5  & 64.7  & -     & -     &\rownum  \\
    & PMM~\cite{pmm}          & ICCV 21       & WRN-38      &           & WRN-38    & ImageNet  & -     & 68.5  & 69.0  & -     & 36.7  &\rownum  \\
    & AMR~\cite{amr}          & AAAI 22       & RN-50       &           & RN-101    & ImageNet  & 69.7  & 68.8  & 69.1  & -     & -     &\rownum  \\
    & SIPE~\cite{sipe}        & CVPR 22       & RN-50       &           & RN-101    & ImageNet  & 68.8  & 68.8  & 69.7  & -     & 40.6  &\rownum  \\
    \cmidrule(r){2-13}
    & RIB~\cite{rib}          & NeurIPS 21    & RN-50       &           & RN-101    & ImageNet  & 70.6  & 68.3  & 68.6  & 45.6  & 44.2  &\rownum  \\
    & ReCAM~\cite{recam}      & CVPR 22       & RN-50       &           & RN-101    & ImageNet  & 70.9  & 68.5  & 68.4  & 44.1  & 42.9  &\rownum  \\
    \cmidrule(r){2-13}
    & OOA~\cite{ooa}          & ICCV 19       & RN-101      &           & RN-101    & ImageNet  & -     & 65.2  & 66.4  & -     & -     &\rownum  \\ 
    & EDAM~\cite{edam}        & CVPR 21       & RN-50       &\checkmark & RN-101    & COCO      & 68.1  & 70.9  & 71.8  & -     & -     &\rownum  \\
    & PMM~\cite{pmm}          & ICCV 21       & WRN-38      &           & WRN-38    & ImageNet  & -     & 68.5  & 69.0  & -     & 36.7  &\rownum  \\
    & URN~\cite{urn}          & AAAI 22       & RN-101      &           & RN-101    & ImageNet  & -     & 69.5  & 69.7  & -     & 40.7  &\rownum  \\
    & ESOL~\cite{esol}        & NeurIPS 22    & RN-50       &           & RN-101    & COCO      & 68.7  & 69.9  & 69.3  & 44.6  & 42.6  &\rownum  \\
    & MCTformer~\cite{mct}    & CVPR 22       & DeiT-S      &           & WRN-38    & VOC       & 69.1  & 71.9  & 71.6  & -     & 42.0  &\rownum  \\
    & TSCD~\cite{tscd}        & AAAI 23       & MiT-B1      &           & MiT-B1    & ImageNet  & -     & 67.3  & 67.5  & -     & 40.1  &\rownum  \\
    \hline
    \multirow{10}*{\rotatebox{90}{Cross-Image}}
    & MCIS~\cite{mcis}        & ECCV 20       & VGG-16      &           & RN-101    & ImageNet  & -     & 66.2  & 66.9  & -     & -     &\rownum  \\
    & CIAN~\cite{cian}        & AAAI 20       & RN-101      &           & RN-101    & ImageNet  & -     & 64.3  & 65.3  & -     & -     &\rownum  \\
    & MBM-Net~\cite{mbmnet}   & ACMMM 20      & RN-50       &           & RN-50     & ImageNet  & -     & 66.2  & 67.1  & -     & -     &\rownum  \\
    & CTI~\cite{cti}          & CVPR 24       & DeiT-S      &           & WRN-38    & -         & 73.7  & 74.1  & 73.2  & -     & 45.4  &\rownum  \\
    \cmidrule(r){2-13}
    & Group-WSSS~\cite{group} & AAAI 21       & RN-101      &           & RN-101    & ImageNet  & -     & 68.2  & 68.5  & -     & -     &\rownum  \\
    & HGNN~\cite{hgnn}        & ACMMM 22      & WRN-38      &\checkmark & RN-101    & COCO      & 70.2  & 70.5  & 71.0  & -     & 34.5  &\rownum  \\ 
    \cmidrule(r){2-13}        
    & SCE~\cite{sce}          & CVPR 20       & WRN-38      &           & RN-101    & ImageNet  & -     & 66.1  & 65.9  & -     & -     &\rownum  \\
    & RCA~\cite{rca}+OOA      & CVPR 22       & WRN-38      &\checkmark & RN-101    & ImageNet  & 73.2  & 71.1  & 71.6  & -     & 35.7  &\rownum  \\
    & RCA~\cite{rca}+EPS      & CVPR 22       & WRN-38      &\checkmark & RN-101    & ImageNet  & 74.1  & 72.2  & 72.8  & -     & 36.8  &\rownum  \\
    & LPCAM~\cite{lpcam}      & CVPR 23       & RN-50       &           & RN-101    & ImageNet  & 71.2  & 68.6  & 68.7  & 46.8  & 44.5  &\rownum  \\
    \hline
    \multirow{4}*{\rotatebox{90}{External}}
    & SSNet~\cite{ssnet}      & ICCV 19       & VGG-16      &\checkmark & VGG-16    & ImageNet  & -     & 63.3  & 64.3  & -     & -     &\rownum  \\
    & EPS~\cite{eps}          & CVPR 21       & WRN-38      &\checkmark & RN-101    & COCO      & 69.4  & 70.9  & 70.8  & -     & -     &\rownum  \\
    \cmidrule(r){2-13}
    & OoD~\cite{ood}+AdvCAM   & CVPR 22       & RN-50       &           & RN-101    & ImageNet  & 72.1  & 69.8  & 69.9  & -     & -     &\rownum  \\   
    \hline
    \multirow{5}*{\rotatebox{90}{FMs}}
    & CLIMS~\cite{clims}      & CVPR 22       & RN-50       &           & RN-101    & COCO      & 70.5  & 70.4  & 70.0  & -     & -     &\rownum  \\
    & CLIP-ES~\cite{clipes}   & CVPR 23       & ViT-B/16    &           & RN-101    & ImageNet  & 75.0  & 71.1  & 71.4  & -     & 45.4  &\rownum  \\
    & WeCLIP~\cite{weclip}    & CVPR 24 & ViT-B/16    &           & ViT-B/16  & CLIP      & -     & 76.4  & 76.2  & -     & 47.1  &\rownum  \\
    \cmidrule(r){2-13}
    & SAM (text input)        & -             & ViT-H/16    &           & RN-101    & ImageNet  & 86.4  & 72.7  & 72.8  & 64.4  & 50.8  &\rownum  \\  
    & SAM (zero-shot)         & -             & ViT-H/16    &           & -         & -         & 78.2  & 74.0  & 73.8  & 54.6  & 54.6  &\rownum  \\
    \hline
    & Fully-Supervised        & -             & -           &           & RN-101    & ImageNet  & 100.0 & 75.3  & 75.6  & 100.0 & 53.5  &\rownum  \\
    \bottomrule
  \end{tabular}}
\end{table*}
}

\subsection{CAM-based vs. SAM-based Methods}
To make a methodological comparison, we compile a summary of important factors such as the venue of the original research publication, the classification backbone used, whether the method utilizes a saliency map, the semantic segmentation backbone deployed, and the source of the pretrained parameters. Table~\ref{table:results} (Rows 1-55) presents the performance of the traditional CAM-based WSSS methods and Table~\ref{table:results} (Rows 59-60) presents the performance of the SAM-based methods on VOC and MS~COCO datasets.

\noindent
\textbf{Quality of pseudo masks.}
Both SAM (text input) and SAM (zero-shot) demonstrate outstanding performance. As shown in Table~\ref{table:results}, on the VOC dataset, SAM (text input) achieves a mIoU score of $86.4\%$ (Row 54), surpassing the state-of-the-art traditional CAM-based method RCA~\cite{rca} (Row 51) by a significant margin of $12.7\%$ (Row 53). Similarly, on the MS COCO dataset, it outperforms the state-of-the-art method LPCAM~\cite{lpcam} by an impressive margin of $17.6\%$ (Row 48). While SAM (zero-shot) (Row 55) may not perform quite as well as SAM (text input), it nevertheless outperforms the state-of-the-art traditional CAM-based methods on both datasets.

\noindent
\textbf{Quality of segmentation masks.}
SAM (text input) surpasses traditional methods and closely approaches the performance of the fully-supervised DeepLabV2~\cite{deeplabv2}. As shown in Table~\ref{table:results}, on the VOC dataset, although SAM (text input) significantly outperforms RCA~\cite{rca} in terms of the quality of pseudo masks, their segmentation performance gap (Row 51 and 59) is only $0.5\%$ on the \texttt{val} and $0\%$ on \texttt{test} set. When compared with a fully-supervised method (considered as the upper bound), SAM (text input) remains competitive with a $2.6\%$ gap (Row 59 and Row 61) on the \texttt{val} set and $2.8\%$ on \texttt{test} set. This suggests that both RCA and SAM (text input) are capable of producing pseudo masks of sufficient quality to train the segmentation network effectively. Hence, further improvements in pseudo mask quality yield only marginal gains in segmentation performance. On the more challenging MS~COCO dataset, we observe a more substantial gap. There is a $5.4\%$ performance difference (Row 46 and 59) between CTI~\cite{cti} and SAM (text input) on the \texttt{val} set, and a $2.7\%$ gap (Row 59 and 61) between SAM (text input) and the fully-supervised method. This could suggest that the increase in complexity and diversity in the MS~COCO dataset makes it harder for WSSS methods to reach the performance level of fully-supervised methods. 

We notice a significant gap (Row 59) between the pseudo mask on \texttt{train} set and the segmentation masks on \texttt{val} set and \texttt{test} set of SAM (text input), which suggests that the performance may be limited by the fully-supervised DeepLabV2~\cite{deeplabv2} segmentation model.
In the zero-shot setting, as shown in Table~\ref{table:results}, the results indicate that SAM (zero-shot) outperforms SAM (text input) (Row 59 and 60) on both the \texttt{val} and \texttt{test} sets across both datasets. This could be attributed to the fact that SAM (zero-shot) does not require training an additional segmentation model. Furthermore, SAM (zero-shot) surpasses the fully-supervised DeepLabV2 (Row 59 and 60) on the challenging MS COCO dataset, highlighting the strong capabilities of the foundation models for WSSS.

\subsection{CLIP-based vs. SAM-based Methods}
Table~\ref{table:results} (Rows 56-58) shows the performance of three CLIP-based methods, and Table~\ref{table:results} (Rows 59-60) presents the performance of the SAM-based methods on VOC and MS~COCO datasets.

\noindent
\textbf{Quality of pseudo masks.}
As shown in Table~\ref{table:results}, on the VOC dataset, SAM (text input) surpasses CLIP-ES~\cite{clipes} by a significant margin of $11.4\%$ (Row 57 and 59). While SAM (zero-shot) (Row 60) does not perform quite as well as SAM (text input), it still outperforms CLIP-ES~\cite{clipes} by $3.2\%$ (Row 57 and 60).

\noindent
\textbf{Quality of segmentation masks.}
On the VOC dataset, WeCLIP~\cite{weclip} outperforms both SAM (text input) and SAM (zero-shot), demonstrating the effectiveness of training a decoder on a fixed CLIP backbone. However, on the more challenging MS~COCO dataset, SAM (text input) and SAM (zero-shot) surpass the CLIP-based methods, highlighting the strong capabilities of the SAM models for WSSS.

\subsection{In-depth Results on SAM-based Methods}
\label{sec:exp_sam}
\begin{figure*}[ht]
\centering
\includegraphics[width=0.99\linewidth]{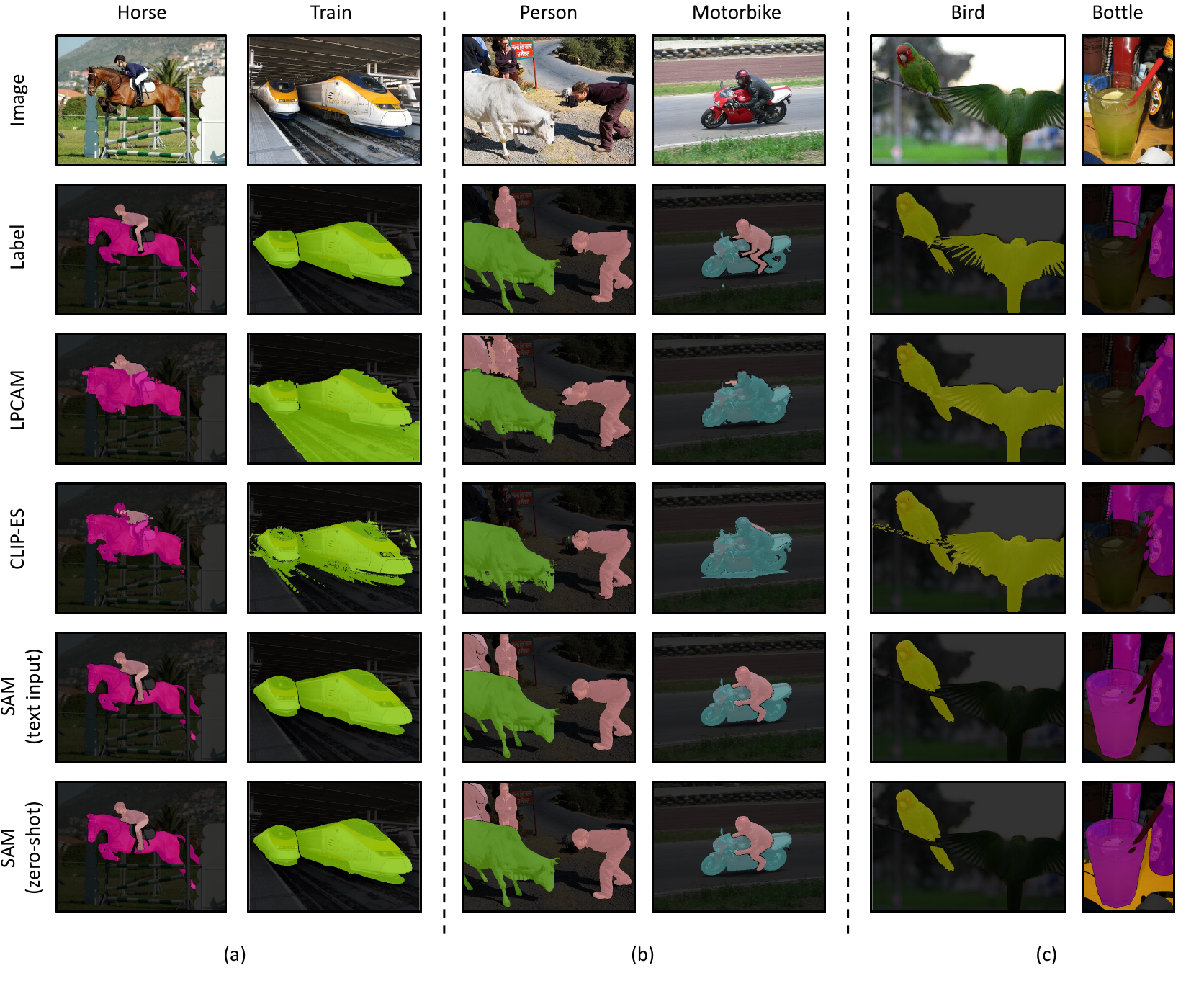}
\caption{Visualization of pseudo masks generated by LPCAM~\cite{lpcam}, CLIP-ES~\cite{clipes}, SAM (text input), and SAM (zero-shot) on VOC dataset. (a) Examples showcasing high-quality masks produced by both SAM (text input) and SAM (zero-shot). (b) Examples where SAM produced masks that even surpass the quality of the ground truth masks. (c) Examples illustrating the failure cases of SAM (text input) and SAM (zero-shot).
}
\Description{
This figure visualizes pseudo masks generated by LPCAM, CLIP-ES, and SAM (evaluated with text input and zero-shot settings) on the VOC dataset. The visualization is divided into three panels: (a) showcases examples (horse and train) where SAM (text input and zero-shot) produces high-quality masks, demonstrating precise object segmentation. (b) highlights cases (person and motorbike) where SAM-generated masks exceed the quality of the ground truth labels, offering more accurate object delineation. (c) presents failure cases (bird and bottle) where SAM (text input and zero-shot) struggles, resulting in inaccurate masks due to misidentified object boundaries or missing details. These comparisons underscore SAM's strengths and limitations across various scenarios.
}
\label{fig:vis}
\end{figure*}

\begin{figure*}[ht]
\centering
\includegraphics[width=0.99\linewidth]{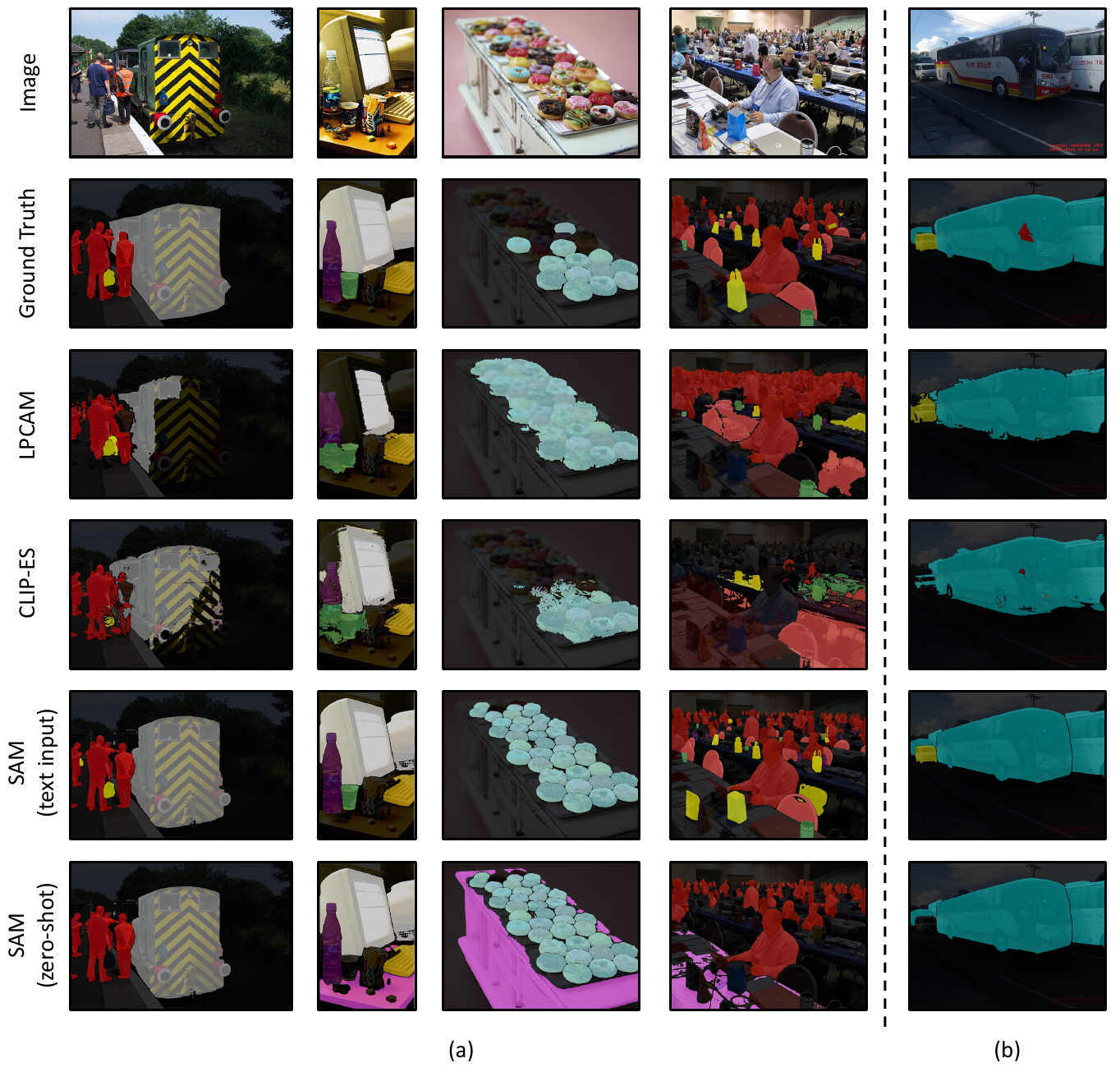}
\caption{Visualization of pseudo masks generated by LPCAM~\cite{lpcam}, CLIP-ES~\cite{clipes}, SAM (text input), and SAM (zero-shot) on MS~COCO dataset. (a) Examples showcasing high-quality masks produced by SAM in complex scenes. (b) Examples illustrating the failure cases of SAM.
}
\Description{
This figure visualizes pseudo masks generated by LPCAM, CLIP-ES, and SAM (evaluated with both text input and zero-shot settings) on the MS COCO dataset. Panel (a) highlights examples where SAM produces high-quality masks, even in complex scenes, effectively segmenting objects with precise boundaries. The rows depict the progression from ground truth masks to outputs from LPCAM, CLIP-ES, SAM (text input), and SAM (zero-shot), showcasing the comparative segmentation quality. Panel (b) illustrates failure cases of SAM, where segmentation results deviate from the ground truth, highlighting challenges in accurately identifying under specific scenarios.
}
\label{fig:vis_coco}
\end{figure*}

\noindent
\textbf{Qualitative results.}
Compared to the traditional method and CLIP-based method, as shown in Fig.~\ref{fig:vis} (a) and (b), both SAM (text input) and SAM (zero-shot) show higher mask quality in terms of the clear boundaries not only between the background and foreground but also among different objects. In the example of ``train'', SAM (text input) and SAM (zero-shot) successfully distinguish between ``train'' and ``railroad'', a challenge that many WSSS methods struggle with due to the co-occurrence of these elements. But SAM (text input) and SAM (zero-shot) will also suffer from the false negative and false positive problem when the tagging model (RAM~\cite{ram}) and grounding model (Grounding DINO~\cite{groundingdino}) wrongly tag or detect some objects. As shown in Fig.~\ref{fig:vis} (c), SAM (text input) fails to identify a ``bird'' and mistakenly classifies a ``cup'' as a ``bottle''. And SAM (zero-shot) wrongly recognizes a ``dining table''. All these failures can be attributed to the grounding model or tagging model rather than SAM, suggesting that the capabilities of the grounding or tagging model could potentially limit the performance of SAM (text input). When compared to the human annotation, Fig.~\ref{fig:vis} (b) highlights cases where SAM (text input) generates masks that even surpass the quality of human annotations. For instance, in the ``person'' example, the human annotation fails to include the person in the top left corner. In the ``motorbike'' example, SAM (text input) can produce precise boundaries in the overlapping scenarios. This suggests the strong capabilities of the foundation models in WSSS.

In the examples of complex scenes in the MS~COCO dataset, as shown in Fig.~\ref{fig:vis_coco} (a), SAM (text input) shows its strong ability to annotate numerous smaller objects (e.g., ``donut'' and ``person''). Similar to the result on the VOC dataset, both SAM (text input) and SAM (zero-shot) also suffer from false negative and false positive problems when the tagging model and grounding model wrongly tag or detect the objects. Fig.~\ref{fig:vis_coco} (b) shows a case where SAM (text input) fails to detect the ``person'' inside the ``bus'', while SAM (zero-shot) overlooks the ``car'' near the ``bus''.  These failures can also be attributed to the grounding model or tagging model rather than SAM.

\setlength{\tabcolsep}{1.6mm}{
\begin{table*}[ht]
\caption{Per-class pseudo mask quality (mIoU) of SAM (text input) and SAM (zero-shot) on VOC dataset. We classify the 20 classes in the VOC dataset into three categories: ``animals'', ``transportation'', and ``objects/items'', with an additional class for ``person''.}
\label{table:per-class}
\centering
\scalebox{0.7}{
\begin{tabular}{lcccccccccccccccccccc}
    \toprule
    && \multicolumn{6}{c}{Animals} & \multicolumn{7}{c}{Transportation} & \multicolumn{6}{c}{Objects/Items} \\
    \cmidrule(r){3-8}\cmidrule(r){9-15}\cmidrule(r){16-21}
    &   \rotatebox{90}{person} &
        \rotatebox{90}{bird} &
        \rotatebox{90}{cat} &
        \rotatebox{90}{cow} &
        \rotatebox{90}{dog} &
        \rotatebox{90}{horse} &
        \rotatebox{90}{sheep} &
        \rotatebox{90}{aeroplane} &
        \rotatebox{90}{bicycle} &
        \rotatebox{90}{boat} &
        \rotatebox{90}{bus} &
        \rotatebox{90}{car} &
        \rotatebox{90}{motorbike} &
        \rotatebox{90}{train} &
        \rotatebox{90}{bottle} &
        \rotatebox{90}{chair} &
        \rotatebox{90}{\parbox{20mm}{dining table}} &
        \rotatebox{90}{\parbox{20mm}{potted plant}} &
        \rotatebox{90}{sofa} &
        \rotatebox{90}{\parbox{20mm}{TV monitor}}   \\
    \cmidrule(r){1-21}
    Text input  & 90.0  & 94.3  & 94.5  & 94.8  & 91.1  & 93.9  & 97.4  & 96.8  & 65.2  & 89.6  & 90.2  & 86.5  & 84.1  & 96.1  & 91.0  & 73.8  & 48.9  & 73.8  & 78.4  & 88.9  \\
    Zero-shot   & 86.0  & 85.9  & 87.0  & 94.8  & 81.4  & 86.7  & 97.3  & 96.6  & 63.9  & 76.9  & 90.1  & 71.0  & 84.7  & 93.9  & 89.2  & 41.1  & 31.2  & 56.7  & 52.5  & 82.9  \\
    \bottomrule
\end{tabular}
}
\end{table*}
}
\noindent
\textbf{Per-class analyses.}
We investigate the per-class pseudo mask quality on the VOC dataset to examine the quality of pseudo masks produced by SAM (text input). As detailed in Table~\ref{table:per-class}, SAM (text input) consistently generates high mIoU scores for most classes related to animals and transportation. However, it fares relatively poorer for classes related to objects or items. We believe that the observed differences could be due to the contrasting nature of the scenes where these subjects are usually found. Specifically, images of animals and transportation modes predominantly portray outdoor scenes, which generally have simpler backgrounds. In contrast, images of objects/items are frequently set indoors and are often characterized by more intricate backgrounds.
Similarly, SAM (zero-shot) displays performance comparable to SAM (text input) in most classes associated with animals and transportation. However, there is a substantial gap when it comes to complex indoor scenes. This implies that the tagging model may encounter challenges in recognizing objects within complex indoor settings.

\begin{figure}[ht]
\centering
\includegraphics[width=0.8\linewidth]{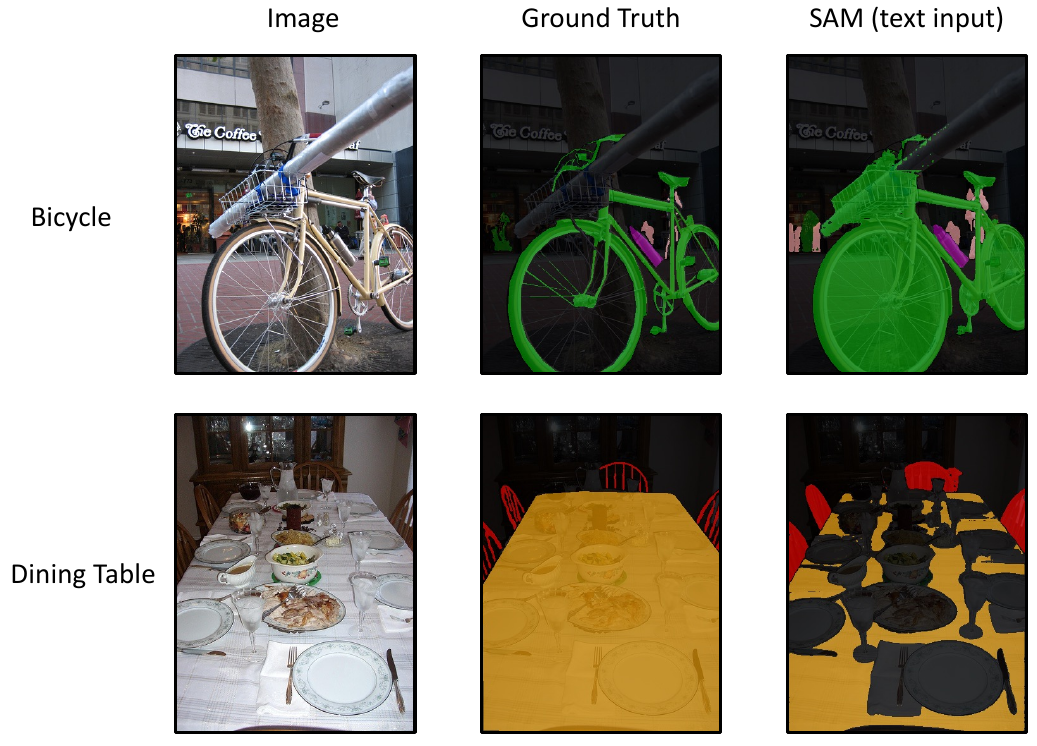}
\caption{Examples (including images, ground truth labels, and pseudo masks produced by SAM (text input)) of bicycles and dining tables on VOC dataset.}
\Description{This figure presents examples from the VOC dataset, featuring bicycles and dining tables. Each example comprises three components: the original image, the corresponding ground truth labels, and the pseudo masks generated by SAM using text input prompts. The pseudo masks delineate object regions identified by SAM, highlighting differences from the ground truth, which arise due to varying annotation protocols.}
\label{fig:bicycle_tables}
\end{figure}
\noindent
\textbf{Failure cases analyses.}
To further delve into the specifics, we focus on the two categories that exhibit the lowest mIoU: bicycles and dining tables. As shown in Fig.~\ref{fig:bicycle_tables}, the peculiarities of annotation protocols in the VOC dataset become evident. The dataset annotations do not include the hollow part of bicycle wheels and do account for items placed on dining tables. However, the results produced by SAM (text input) display the exact opposite tendencies.
Consequently, it becomes clear that this observed deviation is not a flaw inherent to the model. Instead, this discrepancy emerges due to the differences in annotation strategies employed during the creation of the dataset. Hence, it's essential to consider such factors when interpreting the performance of the model across different categories.

\section{Conclusion and Discussion}
\label{sec:conclusion}
In this work, we embark on a deep exploration of weakly supervised semantic segmentation (WSSS) methods, and more specifically, the potential of using foundation models for this task. We first recapitulated the key traditional methods used in WSSS, outlining their principles and limitations. Then, we examine the potential of applying the Segment Anything Model (SAM) in WSSS, in two scenarios: text input and zero-shot. 

Our results demonstrate that SAM (text input) and SAM (zero-shot) significantly outperform traditional methods in terms of pseudo mask quality. In some instances, they even approach or surpass the performance of fully supervised methods. These results highlight the immense potential of foundation models in WSSS. Our qualitative analysis further revealed that the quality of pseudo masks generated by our methods often exceeds the quality of human annotations, demonstrating their capability to produce high-quality segmentation masks. Our in-depth analysis brought to light some of the challenges in applying foundation models to WSSS. One such challenge is the discrepancy between the annotation strategies employed in the creation of the dataset and the models' segmentation approach, leading to divergences in performance. Also, we observed that the complexity of the scenes and the diversity of the categories in datasets could affect the model's performance.

\noindent
\textbf{Limitations and future works.} Despite the considerable success of the SAM-based approach, several limitations remain that present opportunities for future work: 

\begin{itemize}
    \item \textbf{Performance bottleneck.} As discussed in Section~5.4, the performance of our proposed SAM-based methods is constrained by the limitations of the tagging and grounding models (``bird'' and ``bottle'' cases in Fig.~\ref{fig:vis}~(c)), as well as by the annotation protocols (examples in Fig.~\ref{fig:bicycle_tables}). Enhancing the accuracy and reliability of the tagging and grounding models is essential for improving the overall performance of our proposed SAM-based WSSS pipeline. To address the annotation protocol problem, future works can include standardizing annotation protocols to ensure consistent and high-quality mask generation and fine-tuning the model on target datasets to learn the specific annotation protocols.
    \item \textbf{Effectively leveraging SAM in WSSS.} We introduced a zero-shot method leveraging SAM, which shows promise but faces challenges. Recently, methods that integrate SAM into the training process have been proposed. For example, Hyeokjun et~al.~\cite{sam2cam} utilize SAM to segment all objects within a given input image and then compute segment-wise prototype features by averaging the feature maps of each segment. These segment-wise prototypes are subsequently used to perform regional prototype-based contrastive learning. It improves the quality of CAMs by directly transferring SAM's knowledge to the proposed model during training. Another approach~\cite{foundation_wsss} proposes a coarse-to-fine framework based on both CLIP and SAM to generate segmentation seeds for WSSS. This framework employs a frozen SAM image encoder and a frozen CLIP model with learnable task-specific prompts to perform image classification and seed segmentation tasks jointly. Those methods show promising results when SAM is integrated into the proposed workflows. However, these approaches only utilize a frozen SAM during the pseudo-mask generation stage. We believe there is further potential to explore SAM's capabilities during the segmentation stage, given its strong segmentation ability. For example, SAM could be used as a shared backbone for both classification and segmentation, with fine-tuning on the target dataset to enhance performance.
    \item \textbf{Efficiency.} Our proposed SAM-based approaches require multiple large models during inference. The input images will be fed forward to different image encoders multiple times to get the final segmentation result, leading to significant computational overhead and slower inference speeds. Reducing this computational burden is essential for practical deployment. One solution is to reduce the model size. Techniques such as model compression and knowledge distillation can be explored to achieve the goals. For example, MobileSAM~\cite{mobile_sam} distills the knowledge from the heavy image encoder (ViT-H in the original SAM) to a lightweight image encoder. The resulting lightweight SAM is more than 60 times smaller yet performs on par with the original SAM. However, the performance impact of replacing multiple models with lightweight alternatives is not clear. The trade-off needs to be considered for balancing the reduction in computational overhead across the tagging model, grounding model, and SAM itself against the potential decrease in overall performance. 
    \item \textbf{Generalization Ability.} While SAM-based WSSS methods have shown promising results on natural image datasets like VOC and MS~COCO, their performance in other domains, such as medical imaging, remote sensing, and industrial inspection, remains thoroughly evaluated. These domains present unique challenges and opportunities for segmentation. Taking medical imaging as an example, Mazurowski et~al.~\cite{sam_medical} perform an extensive evaluation of SAM’s ability for interactive medical image segmentation on a collection of 19 medical imaging datasets from various modalities and anatomies. They conclude that SAM shows impressive zero-shot segmentation performance for certain medical imaging datasets but moderate to poor performance for others. To enable prompt-free semantic segmentation, SAMed~\cite{samed} applies the low-rank-based fine-tuning strategy to the SAM image encoder and fine-tunes it together with the mask decoder on labeled medical image segmentation datasets. It can achieve highly competitive and remarkable results compared with the previous well-designed medical image segmentation models. Those works demonstrate that SAM's strong segmentation capabilities can be effectively extended to the medical domain. WSSS has been utilized in the medical imaging domain~\cite{wsss_med1,wsss_med2} when the fine-grained annotations are unavailable. Future research could explore SAM-based WSSS methods in the medical imaging domain to address the challenge of data scarcity. In addition to its generalization to other domains, WSSS can also be extended to other tasks by leveraging the pseudo-mask generation techniques developed within WSSS. For instance, in semi-supervised semantic segmentation, a strong-weak dual-branch structure~\cite{ssss} can be employed to mitigate performance degradation caused by the inaccurate supervision of weakly labeled data. Lee et~al.~\cite{wsss_ssss} applies the proposed WSSS pseudo-mask generation method to the weak branch, producing more accurate pseudo-masks for this branch. This approach can significantly enhance the overall performance of semi-supervised semantic segmentation. Exploring how to generalize WSSS to facilitate other tasks is a promising research direction.
\end{itemize}

\section{Acknowledgment}
The author gratefully acknowledges the support from the DSO research grant awarded by DSO National Laboratories, Singapore.


\bibliographystyle{ACM-Reference-Format}
\bibliography{ref}

\end{document}